\documentclass[sigconf]{acmart}

\usepackage{booktabs} % For formal tables
\usepackage{color}
\usepackage[linesnumbered,ruled,vlined]{algorithm2e}
\usepackage{graphicx}
\usepackage{subcaption}
\usepackage{multirow}
\usepackage{fix-cm}
\acmDOI{10.475/123_4}

\acmISBN{123-4567-24-567/08/06}

\begin{document}
\title{Identify Susceptible Locations in Medical Records via Adversarial Attacks on Deep Predictive Models}
% \titlenote{Produces the permission block, and
  % copyright information}
% \subtitle{Extended Abstract}
% \subtitlenote{The full version of the author's guide is available as
%   \texttt{acmart.pdf} document}

\author{Mengying Sun$^1$, Fengyi Tang$^1$ Jinfeng Yi$^2$, Fei Wang$^3$, Jiayu Zhou$^1$}
\affiliation{%
  \institution{$^1$Computer Science and Engineering, Michigan State University, East Lansing, MI, USA\\
               $^2$Tencent AI Lab, Bellevue, WA, USA\\
               $^3$Department of Healthcare Policy and Research, Weill Cornell Medical School, New York, NY, USA
               }
}
\email{{sunmeng, tangfeng}@msu.edu, jinfengyi.ustc@gmail.com, few2001@med.cornell.edu, jiayuz@msu.edu}

\renewcommand{\shortauthors}{M. Sun et al.}
\renewcommand{\shorttitle}{Adversarial Attacks on Medical Deep Predictive Models}

\begin{abstract}
The surging availability of electronic medical records (EHR) leads to
increased research interests in medical predictive modeling. 
Recently many deep learning based predicted models are also developed for EHR data and demonstrated impressive performance. However, a series of recent
studies showed that these deep models are \textit{not safe}: they suffer from certain 
vulnerabilities. In short, a well-trained deep network can be extremely
sensitive to inputs with negligible changes. These inputs are referred to as \textit{adversarial
examples}. In the context of medical informatics, such attacks could alter the
result of a high performance deep predictive model by slightly perturbing a
patient's medical records. Such instability not only reflects the weakness of deep
architectures, more importantly, it offers a guide on detecting susceptible
parts on the inputs. In this paper, we propose an efficient and
effective framework that learns a time-preferential minimum attack targeting
the LSTM model with EHR inputs, and we leverage this attack strategy to screen
medical records of patients and identify susceptible events and measurements.
The efficient screening procedure can assist decision makers to pay
extra attentions to the locations that can cause severe consequence if not
measured correctly. We conduct extensive empirical studies on a real-world
urgent care cohort and demonstrate the effectiveness of the proposed screening approach. 
% {\color{red}Results show that A\% of the patients can be attacked by
% perturbing only B records per person with average perturbation less than C\%
% and maximum perturbation not exceeding D\%, suggesting that additional
% attention should be paid by clinical professionals on those records locations.}

%by leveraging existing algorithms to attack and propose an: 1) features are homogeneous, 2) no temporal structure exists,   and in order to address aforementioned challenges, we propose an effective and efficient optimization-based attack for recurrent neural networks (RNNs) that applied to medical time series data. Our goal is to utilize time-preferential and collaborative attacks to identify sensitive events and measurements in a patient's medical records. Results demonstrate the vulnerability of RNNs and suggest additional attention is needed from domain professionals when necessary.

%Deep networks are shown to behave irregularly under such attacks. 
\end{abstract}




\maketitle

\section{Introduction}\label{sec:intro}
% !TEX root = ./main.tex
Recent years have witnessed substantial success on applying deep learning
techniques in data analysis in various application domains, such as computer vision, natural
language processing, and speech recognition. Those modern machine learning techniques have also demonstrated great potentials in clinical informatics~\cite{shickel2017deep}. For
example, deep learning has been used to learn effective representations for
patient records~\cite{miotto2016deep,choi2016multi} to support disease phenotyping~\cite{che2015deep} and conduct predictive
modeling~\cite{che2016interpretable,pham2016deepcare,nguyen2017mathtt}.
A recent study from Google demonstrated the capability of
deep learning methods on predictive modeling with electronic health records
(EHR) over traditional state-of-art approaches~\cite{Rajkomar2018scalable}.

Deep learning approaches have a few key advantages over traditional machine
learning approaches, including the capability of exploring complicated
relationships within the data through the highly non-linear architecture,
and building an end-to-end analytics pipeline without the process of handcrafted feature engineering. Most of these perks are backed by complex neural
networks and a large volume of training data. However, such complex networks
could lead to vulnerable decision boundaries according to statistical learning
theory. This effect could be further exacerbated by the sparse, noisy and
high-dimensional nature of medical data. For example, in our experiments,
we show that a well-trained deep model may classify a dying patient to be
healthy when the patient's record changed a bit, especially for those close to
the decision boundary. In addition, certain clinical measurements may be more
susceptible to this type of perturbation than others for a given patient. In
this work, we propose to take advantage of such vulnerability of deep
neural networks to identify susceptible events and measurements in each patient's
medical records such that additional attention from clinicians/nurses is required.

The vulnerability of deep neural networks has been brought up in recent
studies, e.g., Szegedy \emph{et al.} first introduced this concept when investigating
the properties of neural networks~\cite{szegedy2013intriguing}. They found
that even a high-performance deep model can be easily ``fooled'', e.g., 
an image classified correctly by the model can be misclassified with human-imperceptible 
perturbations. Plenty of later studies
demonstrated neural networks to be fragile under these so-called
\emph{adversarial attacks}, where \emph{adversarial examples} were generated
to attack deep models using elegantly designed algorithms. Intuitively, if we can attack a high performance medical predictive model and
generate such adversarial medical records from the original medical records
of one patient, then these perturbations in the medical records can inform us
where the susceptible events/measurements are located.

Currently, most existing attack techniques focused on image related
tasks where convolutional neural networks (CNNs) are primarily used. In
the medical informatics domain, however, one major focus remains on predictive
modeling with {\em sequential} medical records~\cite{choi2016multi,zhou2014micro}. In
order to craft efficient and effective adversarial examples, several
challenges persist despite the progress of current attack algorithms. First,
unlike images, medical features are heterogeneous, carrying different
levels and aspects of information, thus resulting in different tolerance to
perturbations. Second, for time sequence data, the effectiveness of
perturbations may vary along time, e.g., perturbing a distant time stamp may
not work as well as a recent one. Third, since we are interested in utilizing attacks to infer susceptible locations, a sparse attack is preferred over a dense one. However, sparse attacks tend to have larger magnitudes than dense attacks, and no explicit evaluation metrics for sparse attacks has been established yet.%based on both criteria yet.

Therefore, to address the aforementioned challenges, we propose an
effective and efficient framework for generating adversarial examples for temporal sequence data. Specifically, our sparse adversarial attack approach is based on optimization and can be efficiently solved via an
iterative procedure, which automatically learns a time-preferential sparse
attack with minimum perturbation on input sequences. From the attack model, we
designed a \emph{Susceptibility Score} for each measurement at both individual-level and
population-level, which can be used to screen medical records from different
patients and identify vulnerable locations. We also define a new evaluation metric that considers both sparsity and magnitude of a certain attack. We evaluate our
attack approach and susceptibility score in the real-world urgent care cohort MIMIC3~\cite{johnson2016mimic},
and demonstrate the effectiveness of the proposed approach from extensive quantitative and qualitative results.  
 In the context of our paper, we mainly verify our methods on medical data, but the attack framework can be easily extended to any other fields.

The rest of the paper is organized as follows: we summarize related work in Section~\ref{sec:related_work}; the proposed framework is presented in Section~\ref{sec:method}; experimental results are shown in Section~\ref{sec:exp} and conclusion reaches at Section~\ref{sec:conclusion}.

\section{Related Work}\label{sec:related_work}
% !TEX root = ./main.tex

Our work lies in the junction of adversarial attacks and recurrent neural networks on medical informatics. Therefore, we briefly summarize recent advances for both fields in this section.

%\subsection{Recurrent Neural Networks on Medical Informatics}
\noindent\textbf{Recurrent Neural Networks on Medical Informatics.}
Recurrent neural network (RNN) and its variants, such as gated recurrent unit
(GRU)~\cite{cho2014learning} and long short-term memory
(LSTM)~\cite{hochreiter1997long}, are designed for analyzing sequence data and
has been widely used in computer vision and natural language processing.
Despite the fact that medical records often consist of time series,
applications of RNN in medical informatics are much less compared to those in
language domain. For example, \cite{lipton2015learning} first utilized LSTM on
EHR for multi-label classification of diagnoses and a similar study
\cite{choi2016doctor} applied GRU to predict diagnose and medication
categories using encounter records. \cite{baytas2017patient} developed 
a time-aware LSTM to address the differences of time intervals in medical 
records. Most recently, a
comprehensive study \cite{Rajkomar2018scalable} showed superior
performance achieved by RNN on predictive modeling with
EHR records compared to traditional approaches, calling for more applications
of such deep models on medical sequential data. While on the other side, the
weakness of deep neural networks was disclosed under more exploration of
network properties.

\noindent\textbf{Adversarial Attacks on Deep Networks.}
Following the discovery of deep network vulnerability, different algorithms
have been developed for crafting adversarial examples to better
understand the robustness of a deep network. The goal of the attacking
algorithm is to crafts an adversarial sample by adding a small perturbation on
a clean sample such that the outcome of a deep model changes after the
perturbation. Below we briefly introduce adversarial attacks on different types of deep models. 

%\paragraph{CNN Attacks}
\textit{CNN Attacks.}
There are many existing studies on CNN attacks. Szegedy \emph{et al.} first
introduced adversarial examples for deep learning in
\cite{szegedy2013intriguing}, where adversarial examples are obtained by
solving an optimization with box-constraints. \cite{goodfellow2014explaining}
proposed a fast gradient sign method (FGS) that uses the gradient of loss
function with respect to input data to generate adversarial examples. 
In DeepFool \cite{moosavi2016deepfool}, an iterative $\ell_2$-regularized algorithm is adopted to find the minimum perturbation that changes the result of a classifier. A universal perturbation \cite{moosavi2016universal} is also formulated later based on DeepFool. JSMA \cite{papernot2016limitations} is a Jacobian-based saliency map algorithm which creates a direct mapping between inputs and outputs during training, and crafts adversarial examples by modifying a fraction of features (the most influential) iteratively. Instead of leveraging network loss and gradients, Carlini and Wagner \cite{carlini2017towards} proposed new objective functions based on the logit layer to generate adversarial examples (C\&W attack). They handle the box-constraint by using a $\tanh$ transformation and also consider different distance metrics ($\ell_0$, $\ell_2$, $\ell_\infty$). 
Chen \emph{et al.} extended C\&W attack to $L_1$ distortion metrics and proposed an elastic-net regularized framework \cite{chen2017ead} for adversarial generation. Zeroth order optimization (ZOO) based attack \cite{chen2017zoo} is a black-box attack algorithm. Different from using network gradients directly, it estimates gradients and Hessian via symmetric difference quotient for crafting adversarial examples. There are other types of adversarial examples like \cite{nguyen2015deep}, and other generating approaches including GAN attacks \cite{zhao2017generating}, ensemble attacks \cite{liu2016delving}, ground truth attacks \cite{carlini2017ground}, hot/cold attacks \cite{rozsa2016adversarial}, feature adversary \cite{sabour2015adversarial} which we do not present further details.

\textit{RNN Attacks.}
Most previous efforts for crafting adversarial examples were made on image classification tasks in the domain of computer vision. Adversarial on deep sequential models are less frequent compared to CNN attacks. For RNN attacks, one focus has been emerged on natural language processing where adversarial examples are generated by adding, removing, or changing words in a sentence \cite{jia2017adversarial, papernot2016crafting, li2016understanding}. However, those perturbations are usually perceptible to human beings. Another application is on malware detection which is a classification task on sequential inputs. \cite{grosse2017adversarial} generated adversarial examples by leveraging algorithms of DNN attacks. \cite{hu2017generating, anderson2016deepdga} used GAN based algorithm and \cite{anderson2017evading} used reinforcement learning to generate adversarial examples. 

In summary, neither have the aforementioned attacks been used to verify the robustness of RNN models on medical sequence data, nor have they been utilized to provide additional important information located in medical records and thus improves the quality of modern clinical care. 

%\paragraph{DNN Defenses}
%An intuitive way for defending against adversarial examples and improving network robustness is to include them in the training phase, known as \textit{data augmentation}, which has been proposed and demonstrated in \cite{goodfellow2014explaining, zheng2016improving, huang2015learning}. \textit{Network distillation} \cite{papernot2016distillation} decreases the effectiveness of adversarial examples by retraining a network which includes the class probabilities (soft labels) as network input additional to training data.  \textit{Adversarial Detecting} tries to distinguish between clean and adversarial examples during the test phase and has been proposed in \cite{gong2017adversarial, lu2017safetynet, metzen2017detecting, feinman2017detecting, meng2017magnet,  hendrycks2017early}. Transforming adversarial examples back to clean input is another way of defense. \cite{gu2014towards} proposed a contractive autoencoder for two-way mapping and \cite{song2017pixeldefend} convert adversarial examples back to training distribution. Other defenses includes network verification \cite{katz2017reluplex, gopinath2017deepsafe} and ensemble defenses \cite{meng2017magnet}.
%
%No defenses have been particularly presented for RNN attacks, but approaches in regular DNN defenses like data augmentation, adversarial detecting are definitely valid to defend against adversarial examples.

\section{Method}\label{sec:method}
% !TEX root = ./main.tex

Though much of the prior efforts have focused on attack and protect strategies
of deep models, in this work, we take a radically different perspective from
existing works and leverage the vulnerability of the deep models to inspect
the features and data points in the datasets that are sensitive to the complex
decision hyperplanes of powerful deep models. When dealing with electronic
health records (EHR), such susceptible locations allows us to develop an
efficient screening technique for EHR. In this section, we first introduce the
problem setting of adversarial attacks, and then propose a novel attack
strategy to efficiently identifying susceptible locations. We then use the
attacking strategy to derive a susceptibility score which can be deployed in healthcare
systems. 

\begin{figure}[t!]
    \centering
    \vspace{-1em}
    \includegraphics[width=\columnwidth]{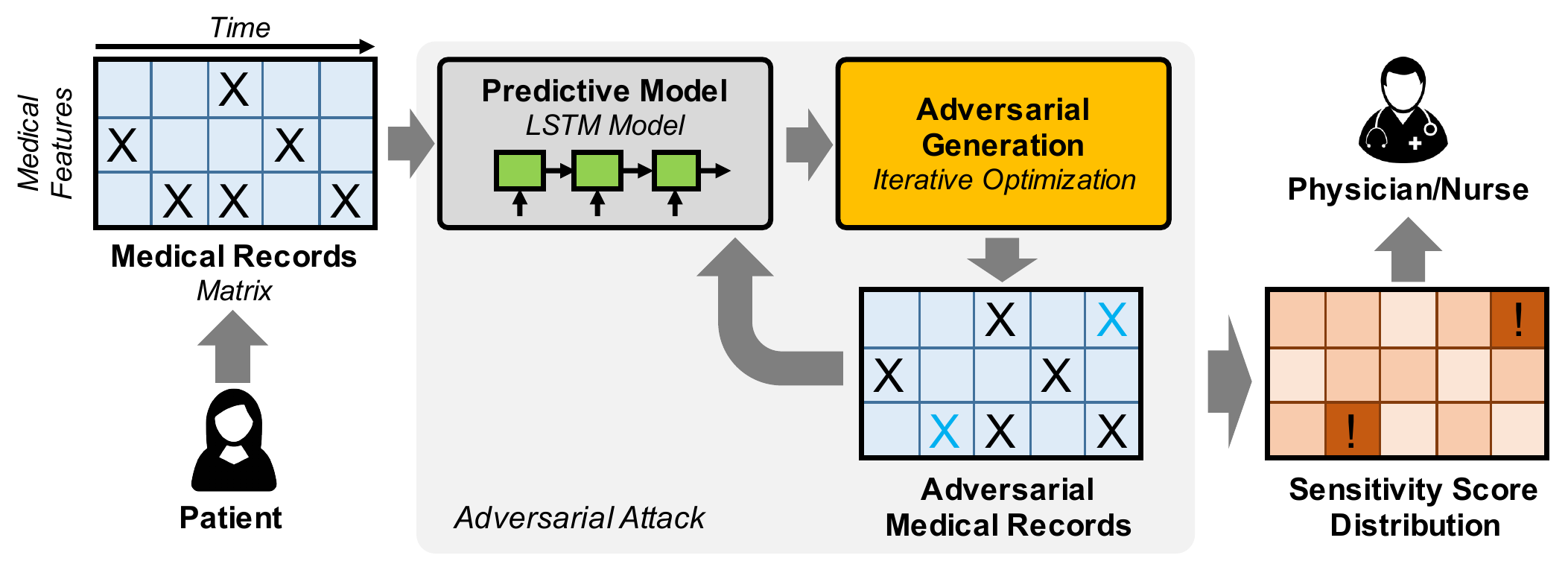}
    \vspace{-2em}
    \caption{Illustration of the proposed framework of identifying sensitive
    locations in electronic medical records. Adversarial records are generated
    by an adversarial attack procedure, which are then used to compute a
    susceptibility score distribution over the medical records. The distribution
    is then used to bring attention to clinicians of entires that cause
    high damage if not accurately recorded or measured.}
    \vspace{-1em}
    \label{fig:framework}
\end{figure}

The proposed framework of adversarial generation is illustrated in Figure
\ref{fig:framework}. It consists of three parts: (1) building a predictive model which
maps time-series EHR data to clinical labels such as diagnoses or
mortality, (2) generating adversarial medical records based on the output of the
predictive model, and (3) computing the susceptibility score based on the adversarial 
samples. 

\subsection{Predictive Modeling from Electronic Medical Records}
Medical records for one patient can be represented by a multivariate time
series matrix~\cite{baytas2017patient,wang2012towards,zhou2014micro}. Assume
we have a set of $d$ medical features in the EHR system and a total of $t_i$
time points available patient $i$, then the patient EHR data can be
represented by a matrix $X^{(i)} \in \mathbb{R}^{d \times t_i} = [x^{(i)}_1,
x^{(i)}_2, \dots , x^{(i)}_{t_i} ]$. Note that for different patients, the
observation length $t_i$ could be different due to the frequency of visits and
length of enrollment. At the predictive modeling step, a model is trained to map EHR features of a patient
to clinically meaningful labels such as diagnoses, mortality and other
clinical assessments. In this paper, we limit our discussions in the scope of
classification, i.e., we would like to learn a $c$-class prediction model
$f(X^{(i)}): \mathbb{R}^{d \times t_i} \rightarrow \{0, 1, \dots,c\}$ from our data. We use
\emph{in-hospital mortality} as the running example, where $0$ and $1$ represent alive and deceased status, respectively. We note that the technique discussed in this
paper can be applied to other predictive tasks such as phenotyping and diagnostic
prediction.

Recurrent neural networks (RNN) are recently adopted in many predictive
modeling studies and achieved state-of-the-art predictive performance, because 
(1) RNN can naturally handle input data of different lengths and (2)
some variants such as long short-term memory (LSTM) networks have demonstrated
excellent capability of learning long-term dependencies and are capable of
dealing with irregular time interval between time series
events~\cite{baytas2017patient}. Without loss of generality, we use a standard
LSTM network as our base predictive model. Given a dataset of $n$ patients,
$\mathcal D = \{(X^{(i)}, y^{(i)})\}_{i=1}^n$, where for each patient we have
a set of medical records $X^{(i)}$ and a target label $y^{(i)}$. The LSTM
network parameters are collected denoted by $\theta$, and the deep network model is
trained by minimizing the following loss function:
\begin{align*}
    \min_{\theta} 
      %J(\mathcal D; \theta)=
      \frac{1}{n} \sum_{i=1}^{n} 
      -\left(
         y^{(i)}\log{\left(y^{(i)}_\theta\right)}+
         \left(1-y^{(i)}\right)\log{\left(1-y^{(i)}_\theta\right)}
       \right),
\end{align*}
where $y^{(i)}_\theta = f(X^{(i)}; \theta)$ is output of the network.
The loss function is minimized by gradient descent via backpropagation
algorithm. Let $\theta^*$ be the parameter obtained by the algorithm
from the training dataset $\mathcal D$, 
and we denote the final predictive model by $f_{\theta^*}(\cdot)$.

\subsection{Adversarial Medical Records Generation}
Our framework is to utilize adversarial attacks to detect susceptible locations
in medical records, and an efficient adversarial attack strategy is the key
component in this framework. 
Adversarial examples are usually generated by exploiting internal state information
of a high performing pre-trained deep predictive model, which can be the intermediate output or gradients. The
existing attack algorithms can be roughly grouped into two prototypes:
\emph{iterative attacks} and \emph{optimization-based attacks}. The former
iteratively add perturbations to original input until some conditions are met.
For example, fast gradient sign method (FGS) \cite{goodfellow2014explaining}
uses gradients of the loss function with respect to input data points to generate
adversarial examples.%  via following equation:
% \begin{align}
% x' = x + \epsilon \text{sign} (\nabla J(x, l')) 
% \end{align}
% where $x$ and $x'$ denotes clean and adversarial data vectors, $l'$ is the desired adversarial target and $\epsilon$ controls the scale of distortion. 
On the other hand, optimization-based attacks cast the
generation procedure into an optimization problem in which the perturbation
can be analytically computed. Optimization based attacks are shown to have
superior performance on adversarial attacks compared to iterative
methods~\cite{chen2017ead, chen2017zoo}.

In order to generate efficient perturbations for medical records, in this
paper we propose an optimization-based attack strategy. Existing attack
strategies typically seek a dense perturbation of an input. Jointly perturbing on
multiple locations simultaneously can take full advantage of the complexity of
the decision surface to achieve minimal perturbation. As long as the magnitude of the perturbed locations is small enough, then it can hardly be perceived by
human. However, such dense
perturbation strategy is not so meaningful in healthcare, as dense perturbations 
could easily change the underlying structures of medical records, and 
introduce comorbidity related to diseases originally not associated with the patient.  
For example, if a perturbation
simultaneously adds the diagnosis features of \emph{Hypertension}
and \emph{Heart Failure} to a patient who never had these before, it is likely
that we are creating new pathology associated to cardiovascular disease. This
suggests that \textit{focused} (sparse) attacks are preferred in our problem. 

Formally, given a well performed predictive model, $f_{\theta^*}(\cdot)$ and a
patient record $X$, our goal is to find an adversarial medical record $\tilde
X$ of the same size, such that $\tilde X$ is close to $X$ but with a different
classification result from the given deep model. Let $y_\theta$ be a
\emph{source label} currently outputted by the predictive model
$f_{\theta^*}(\cdot)$, and we would like the predictive model to classify
$\tilde X$ into an adversarial class label $\tilde y_\theta \neq y_\theta$,
while minimizing the difference $X - \tilde X$. As in our mortality prediction
example, if a patient is originally predicted to be alive, we would like
to find the minimal sparse perturbation $X - \tilde X$ to make our model
predict the deceased label. We note that it is not necessary for this patient to be a
part of the dataset training the model $f_{\theta^*}$. We propose
to obtain the adversarial sample $\tilde X$ by solving the following 
sparsity-regularized attack objective:
\begin{align}
    \min_{\tilde X} 
        \max \left\{ 
           \left[\textit{Logit}(\tilde X)\right]_{y_\theta} 
              - \left[\textit{Logit}(\tilde X)\right]_{\tilde y_\theta}, -\kappa \right\} 
           +\lambda \|\tilde X -X \|_1, \label{eqt:obj}
\end{align}
where \textit{Logit($\cdot$)} denotes outputs before the \textsc{Softmax} layer in the
pre-trained predictive model, $\kappa \geq 0$ ensures a gap between the source
label $y_\theta$ and the adversarial label $\tilde y_\theta$, and $\| A \|_1 =
\sum_{i}\sum{j}|A_{ij}|$ is an element-wise $\ell_1$ norm. To create
adversarial examples with smaller perturbations, $\kappa$ is commonly set to
0. The loss function is similar to the one in C\&W
\cite{carlini2017towards} and EAD attack \cite{chen2017ead}, aiming to assign
label $\tilde y_\theta$ the most probable class for $X$. The $\ell_1$ norm
regularization induces sparsity on the perturbation and encourages the desired focused attacks. Additionally, based on the
hypothesis that different features have different tolerances and temporal
structures might also effect the distribution of susceptible regions, regularization also avoids large perturbations and leads to attacks that have
unique structures at both time and feature levels.

Similar to \cite{chen2017ead}, we use the iterative soft thresholding
algorithm (ISTA) optimization procedure~\cite{beck2009fast} to solve the
objective, where for each record value, the algorithm performs a soft
thresholding $S_{\lambda}(\cdot)$ to shrink a perturbation to 0 if the
deviance to original record is less than $\lambda$ at each iteration, where
soft thresholding performs an element-wise shrinkage of $a$, i.e.,
$S_{\lambda}(a) =
\max(a - \lambda, 0)$. The generation procedure is summarized in
Algorithm~\ref{algo:generator}. 

\begin{algorithm}[tb]
    \DontPrintSemicolon % Some LaTeX compilers require you to use \dontprintsemicolon instead
    \KwIn{A high performance predictive model $f_{\theta^*}(\cdot)$, original clean record $X$ and an adversarial label $\tilde y$, step size $\alpha$, $L_1$ regularization parameter $\lambda$, and the maximum iteration $t_{\max}$.}
    \KwOut{The adversarial record $\tilde X$}
    Initialize $\tilde X \gets X$\;
    \For{$i = 0$ \textbf{to} $t_{\max}$} {
        $\tilde X^{(i+1)}=S_{\lambda}\left(\tilde X^{(i)}-\alpha \cdot \nabla J\left(\tilde X^{(i)}\right)\right),$\\
        \qquad where  $\nabla J(\cdot)$ is the gradient of the loss function in \eqref{eqt:obj}\\
        $\tilde y^{(i+1)}_\theta=f_{\theta^*}\left(\tilde X^{(i+1)}\right)$\\
        \If{$y^{(i+1)}_\theta = \tilde y $} {
            break\;
        }
    }
    \Return{$X'$}\;
    \caption{Adversarial records generation.}
    \label{algo:generator}
\end{algorithm}

Given different strengths of regularization, we end up with a set of adversarial candidates for each medical record matrix. We pick \emph{optimal} adversarial records based on evaluation metrics below.

\subsection{Evaluation of Focused EHR Attacks} 
Recall that in the detection of susceptible locations in medical records, a
\textit{focused} sparse attack is preferred over a dense one due to the
aforementioned reasons. However, using existing attack evaluation metrics such as
the magnitude of the perturbation and accuracy of the attacks, a dense one is almost always more efficient
than a sparse one because the attacks strategies can fully leverage the
complicated decision surface. An analogy is to consider the attacking
strategies in the context of the machine learning paradigm. If we perform
learning and evaluation on the same training data, then the ``best performing''
model will be the one that uses no regularization at all, and therefore the testing data
is critical for a fair evaluation of the learned model. This is why we need an
evaluation strategy that is designed specifically for our attack budget (sparsity).

However, there are currently no metrics in the literature that evaluates both
perturbation scale and degree of focus. Therefore, we propose a novel
evaluation scheme which measures the quality of a perturbation by considering
both the perturbation magnitude and the structure of the attacks.
Given an adversarial record $\tilde X \in \mathbb{R}^{d \times t}$, the perturbation is defined as:
\begin{align*}
    \Delta X = \tilde X - X.
\end{align*}
Thus, the maximum absolute perturbation for an observation ($MAP$) and the percentage of record values that being perturbed ($PP$) can be written as:
\begin{align}
    \text{MAP}(\Delta X) = \max{(|\Delta X|)}, \text{PP}(\Delta X) = \frac{\|\Delta X\|_0}{d\cdot t}, \label{eq: map-pp}
\end{align}
where $|\cdot|$ is element-wise absolute value. To make perturbations comparable, we normalize our data into [0, 1] range using min-max normalization before any experiments (will be covered in section 4). Since both measures behave like a rate ranging from 0 to 1 and we want both to be small, we define the \textit{Perturbation Distance Score} metrics as follows:
\begin{align}
D_{\Delta X} = \sqrt{\text{MAP}(\Delta X)^2 + \beta \cdot \text{PP}(\Delta X)^2},
\end{align}
which geometrically measures a weighted distance between point ($MAP$, $PP$) and original coordinate. $\beta$ is a weighting parameter that controls which measure is emphasized. In our case, we are more concerned about sparsity and set $\beta=2$. The perturbation distance score is used for selecting the best adversarial record for a given observation under different sparsity control. A lower score indicates better quality of perturbation in terms of magnitude and degree of focus. 

\subsection{Susceptibility Scores for EHR Screening} 
Besides individual-level perturbation measure, we are also interested in the
susceptibility of a certain feature or time stamp at a population-level. These 
measurements serve to identify the susceptible locations over the entirety of our medical records.  
We propose three metrics for assessing susceptibility. For simplicity, we assume that the EHR records of all patients have the same length $T$, and we collectively denote the optimal adversarial examples for all patients in a tensor $\tilde{\mathcal{X}} = [\tilde X^{(1)}, \tilde X^{(2)}, \dots, \tilde X^{(n)}] \in \mathbb{R}^{n\times d\times t}$, whose perturbation is denoted by $\Delta\mathcal{X}$. 
For the time-feature grid of time stamp $i$ and feature $j$, we calculate global maximum perturbation ($GMP$), global average perturbation ($GAP$) and the probability of being perturbed across all records ($GPP$):
\allowdisplaybreaks
\begin{align*}
    \text{GMP}_{i, j} &= \max_{1 \le k \le n}{(| \Delta\mathcal{X}_{i, j, k}|)},\\
    \text{GAP}_{i, j} &= \frac{1}{n}\sum\nolimits_{k=1}^{n}{(|\Delta\mathcal{X}_{i, j, k}|)},\\
    \text{GPP}_{i, j} &= \frac{\|\Delta X_{i, j, \cdot}\|_0}{n}.
\end{align*}
The \emph{susceptibility score} for a certain time-feature location is thus defined by:
\begin{align}
S_{ij} = \{\text{GMP} \odot \text{GPP}\}_{ij},
\end{align}
where $\odot$ denotes element-wise product. The rationale behind this score is the expectation of maximum perturbation for a certain time-measurement grid considering all observations within a population. A larger score indicates high susceptibility of corresponding location with respect to certain diagnose under current predictive model. We also adopt a \emph{cumulative susceptibility score} for each measurement defined as:
\begin{align}
S_j = \sum\nolimits_{i=1}^{t} \{\text{GMP} \odot \text{GPP}\}_{ij},
\end{align}
which indicates overall susceptibility at the measurement level.
 
\subsection{Screening Procedure}

We summarize here the procedure of the proposed susceptibility screening
framework. For each medical record, we utilize the attack algorithm
described in Algorithm~\ref{algo:generator} to perturb the existing deep predictive model until the classification result changes. At each iteration, we evaluate on
current adversarial record. Once the result changes, we stop and output the result
which can be used for calculating different scores. We picked the optimal
adversarial example for each EHR matrix based on perturbation distance score.
We then use the adversarial example to compute the susceptibility score for the
EHR as well as the cumulative susceptibility score for different measurements.

\section{Experiment}\label{sec:exp}
% !TEX root = ./main.tex
\subsection{Data Preprocessing}
In this work, we use MIMIC-III (Medical Information Mart for Intensive Care III) \cite{johnson2016mimic} as our primary data. This dataset contains health-related information for over 45,000 de-identified patients who stayed in the critical care units of the Beth Israel Deaconess Medical Center between 2001 and 2012. MIMIC-III contains information about patient demographics, hourly vital sign measurements, laboratory test results, procedures, medications, ICD-9 codes, caregiver notes and imaging reports.

Our experiment uses records from a collection of patients, each being a multivariate time series consisting of 19 variables from vital sign measurements (6) and lab events (13). Vital signs include heart rate (HR), systolic blood pressure (SBP), diastolic blood pressure (DBP), temperature (TEMP), respiratory rate (RR), and oxygen saturation (SPO2). Lab measurements include: Lactate, partial pressure of carbon dioxide (PaCO2), PH, Albumin (Alb), HCO3, calcium (Ca), creatinine (Cre), glucose (Glc), magnesium (Mg), potassium (K), sodium (Na), blood urea nitrogen (BUN), and Platelet count.

\noindent\textbf{Imputation.} MIMIC-III contains numerous missing values and outliers. %We impute both to maintain the most information. <-- did you mean impute missing values and remove outliers?
 We first impute missing values using average across time stamps for each record sequence. Then we remove and impute outlier recordings using the interquartile range (IQR) criteria. For each feature, we flatten across all subjects and time stamps and calculate the IQR value. Lower and upper bound values are defined as 1.5 IQR below the first quartile ($Q_1$) and 1.5 IQR above the third quartile ($Q_3$). Values out of these bounds are considered as outliers. For each outlier, we impute its value in a carry-forward fashion from the previously available time stamp. If the outlier occurs at the first time stamp, we impute its value using the EHR average across all remaining time stamps.

\noindent\textbf{Padding.} One challenge facing EHR data is that time-series recordings are measured asynchronously, often yielding sequential data of different lengths. In previous works, this problem was addressed by taking hourly averages of each feature over the course of an admission ~\cite{harutyunyan2017multitask}. This method allows for hourly alignment features for each patient across visits. For our temporal features, the mean length of observation is 60, and median length is 32. We pad all sequences into the same length (48 hours) by pre-padding short sequence using a masking value and pre-truncating long sequences. Therefore all the sequences are aligned to the most recent events and are aligned across each time step.

\noindent\textbf{Normalization.} After imputation and padding, we obtain a tensor with three dimensions: observation, time and feature. We normalize using min-max normalization, i.e., for each feature, we collect minimum and maximum values across all observations and time stamps and normalize by:
\begin{align}
    X_{\text{new}} = \frac{X-\min{(X)}}{\max{(X)}-\min{(X)}},
\end{align}
where the reason is that different measurements have different range and scale, and possibly different tolerance to perturbations. We want to make perturbations among measurements comparable and the calculation of perturbation score more consistent. After normalization, we have 37,559 multivariate time series with 19 variables across 48 time stamps.

\begin{table}[b!]
    \vspace{-0.15in}
    \centering
    \caption{Performance of 5-fold cross validation for LSTM.}
    \label{lstm-result}\Small
    \vspace{-0.15in}
    % \begin{tabular}{|c|c|c|}
    %     \hline
    %     Metric    & Average & SD     \\ \hline
    %     AUC       & 0.9094  & 0.0053 \\ \hline
    %     f1        & 0.5429  & 0.0194 \\ \hline
    %     Precision & 0.4100  & 0.0272 \\ \hline
    %     Recall    & 0.8071  & 0.0269 \\ \hline
    % \end{tabular}
    \begin{tabular}{c | c c c c}
        \hline
        Metric         & AUC 
                       & F1 
                       & Precision 
                       & Recall\\ \hline
        Avg.$\pm$SD & 0.9094$\pm$0.0053
                       & 0.5429$\pm$0.0194
                       & 0.4100$\pm$0.0272
                       & 0.8071$\pm$0.0269\\ \hline
    \end{tabular}
    \vspace{-1em}
\end{table}

\subsection{Predictive Modeling Performance}

\noindent\textbf{Imbalanced data.} In order to train a good classifier, we use 5-fold cross-validation. However, among 37,559 observations, there are only 11.1\% (4153) deceased labels which make our data highly imbalanced. Therefore, we down-sample observations from the negative class during train-test splitting. We split data into 5 folds in a stratified manner so that the ratio of positive to negative class remains the same as the original dataset on each split. With 5-fold validation, each fold serves as testing fold once, and the other 4 folds become merged and down-sampled to produce balanced training sets. We split the balanced training fold again by 4:1 ratio and treat as training and validation set respectively. 

\noindent\textbf{Network Architecture.} Considering that an overtly complex architecture may increase the instability of the network, we use a similar architecture as \cite{lipton2015learning} but with simpler settings. Our network contains 3 layers: an LSTM layer, a fully-connected layer followed by soft-max layer. We choose the number of nodes using cross-validation and end up with each layer containing 128, 32 and 2 hidden nodes respectively. We then retrain the entire training folds (training + validation) and assess performance on the testing set, which can be found in Table \ref{lstm-result}. The resulting LSTM classifier achieves average 0.9094 (0.0053) AUC and 0.5429 (0.0194) f1 score across all folds.

%\begin{table}[!h]
%    \centering
%    \caption{Performance of 5-fold cross validation for LSTM.}
%    \label{lstm-result}
%    \begin{tabular}{|c|c|c|c|c|}
%        \hline
%        Fold               & AUC    & f1     & Precision & Recall \\ \hline
%        1                  & 0.9064 & 0.5374 & 0.3996    & 0.8200 \\ \hline
%        2                  & 0.9024 & 0.5331 & 0.4065    & 0.7742 \\ \hline
%        3                  & 0.9150 & 0.5315 & 0.3919    & 0.8259 \\ \hline
%        4                  & 0.9142 & 0.5351 & 0.3942    & 0.8331 \\ \hline
%        5                  & 0.9091 & 0.5774 & 0.4576    & 0.7823 \\ \hline
%        Average            & 0.9094 & 0.5429 & 0.4100    & 0.8071 \\ \hline
%        Standard Deviation & 0.0053 & 0.0194 & 0.0272    & 0.0269 \\ \hline
%    \end{tabular}
%\end{table}

\subsection{Susceptibility Detection}
We apply the proposed attack framework on correct classified samples and generate susceptibility scores at two levels: patient level and cohort level.

\begin{figure}[t!]
    \vspace{-0.2in}
    \centering
    \begin{subfigure}[b]{0.25\textwidth}
        \includegraphics[width=\textwidth]{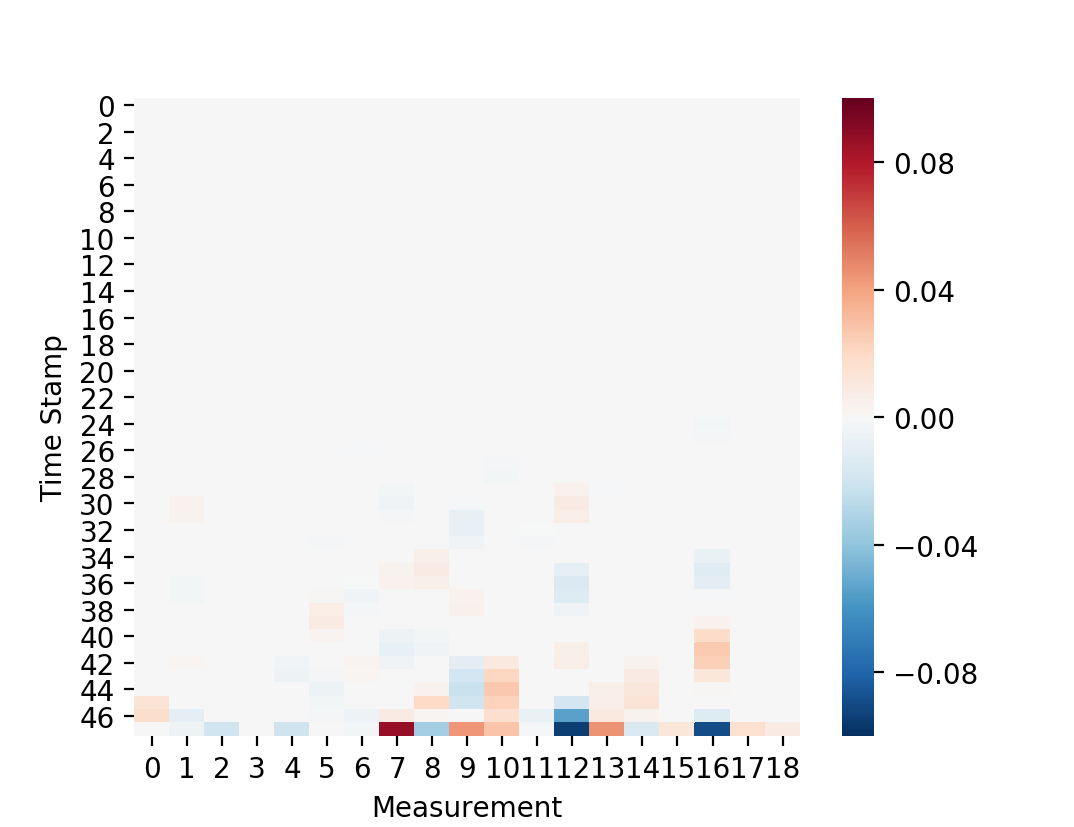}
        \caption{Perturbation 0 to 1.}
        \label{fig:p-01a}
    \end{subfigure}
    ~ %add desired spacing between images, e. g. ~, \quad, \qquad, \hfill etc. 
    \begin{subfigure}[b]{0.21\textwidth}
        \includegraphics[width=\textwidth]{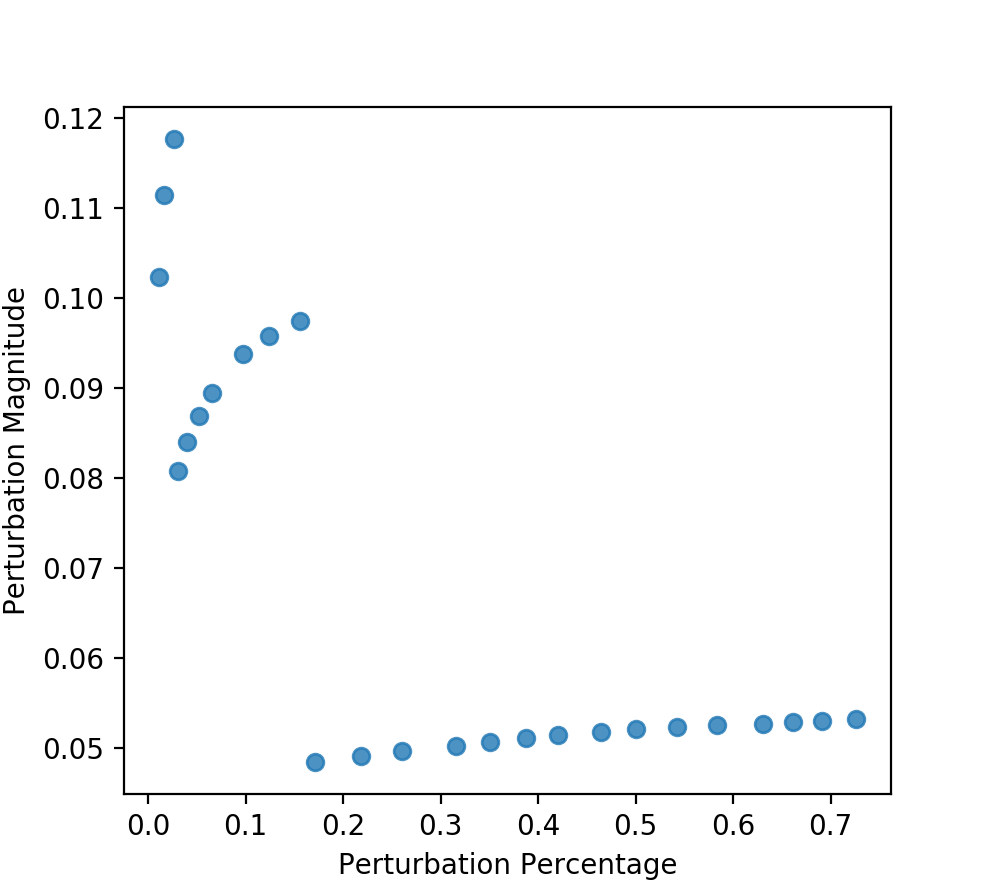}
        \caption{Magnitude vs Sparsity.}
        \label{fig:p-01b}
    \end{subfigure}
    
    \begin{subfigure}[b]{0.24\textwidth}
        \includegraphics[width=\textwidth]{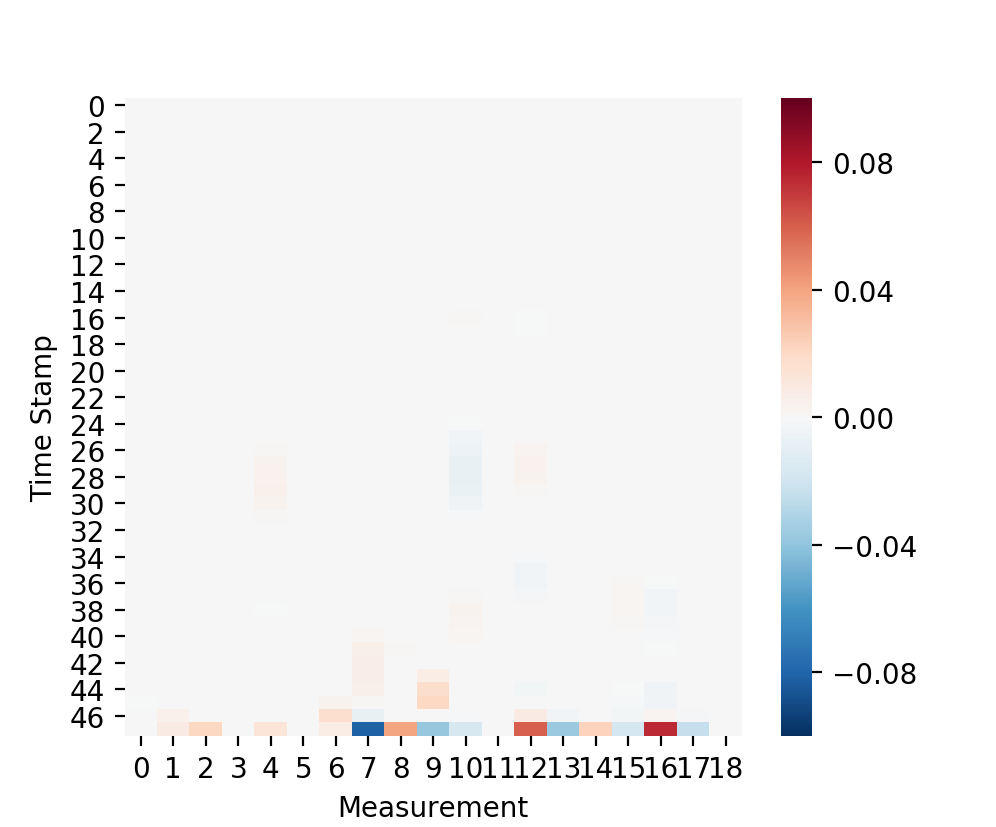}
        \caption{Perturbation 1 to 0.}
        \label{fig:p-10a}
    \end{subfigure}
    ~ %add desired spacing between images, e. g. ~, \quad, \qquad, \hfill etc. 
    \begin{subfigure}[b]{0.22\textwidth}
        \includegraphics[width=\textwidth]{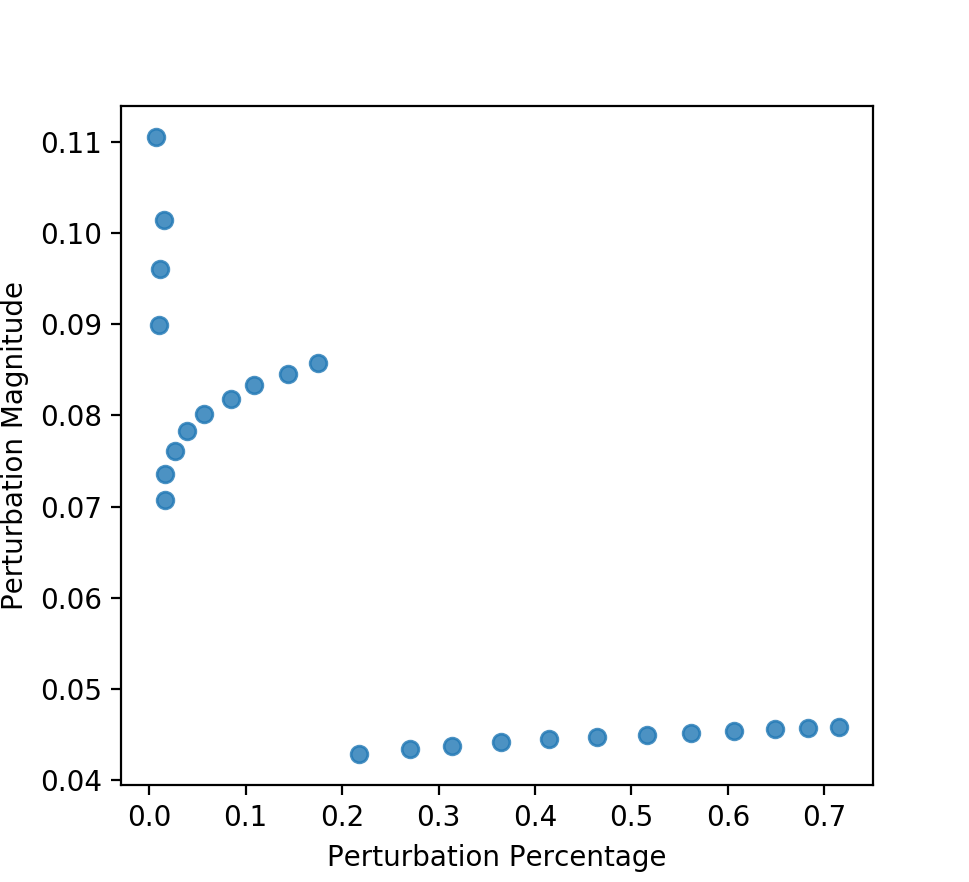}
        \caption{Magnitude vs Sparsity.}
        \label{fig:p-10b}
    \end{subfigure}
    \vspace{-0.1in}
    \caption{Adversarial attack for random chosen patients.}\label{fig:patient-attack}
    \vspace{-0.3in}
\end{figure}

\vspace{+0.5em}
\noindent\textbf{Sensitivity at patient Level.}
There exists two directions of attacks in binary classification tasks, ie., perturb from negative class to positive and vice versa. Figure \ref{fig:patient-attack} shows the adversarial attack results for two patients from each of the attack groups. Figure \ref{fig:p-01a} and \ref{fig:p-10a} illustrates a distribution of perturbations for a successful adversarial record, ie., $X'-X$. The color represents perturbation magnitude at each measurement and time stamp. Note that the adversarial records which generate the perturbations are chosen from a series of candidate adversarial records via minimizing the perturbation distance score mentioned previously. 

We can see that most of the spots are zeros, indicating that no perturbations are made. Among the perturbed locations, we observe a clear pattern along the time axis, which indicates that recent events are more likely to be perturbed compared to distant events. This is consistent with our initial hypothesis that structured perturbation can be more effective. In fact, due to the remember-and-forget mechanism of LSTM networks, the attack algorithm automatically learns to attack this type of model with different degrees of focus for different locations. Our second hypothesis is also verified by the fact that different measurements tend to have varying tolerance to perturbations, as some measurements are not chosen to be attacked while others are more frequently perturbed. 

Figure \ref{fig:p-01b} and \ref{fig:p-10b} presents the maximum perturbation for a patient under different sparsity control. Each regularization parameter generates an adversarial candidate. We can observe the trade-off between perturbation magnitude and sparsity. When only a few spots are attacked, in order to generate success attack, the magnitude of perturbation tends to be large; As more locations change, the perturbation can be flattened to each spot and the magnitude drops quickly. Once over a certain threshold, the maximum perturbation remains similar afterward.

\begin{figure}[t!]
    \vspace{-0.1in}
    \centering
    \begin{tabular}{c  c}
        \textbf{(0 $\Rightarrow$ 1) Attack} & \textbf{(1 $\Rightarrow$ 0) Attack}\\ 
        \includegraphics[width=0.26\textwidth]{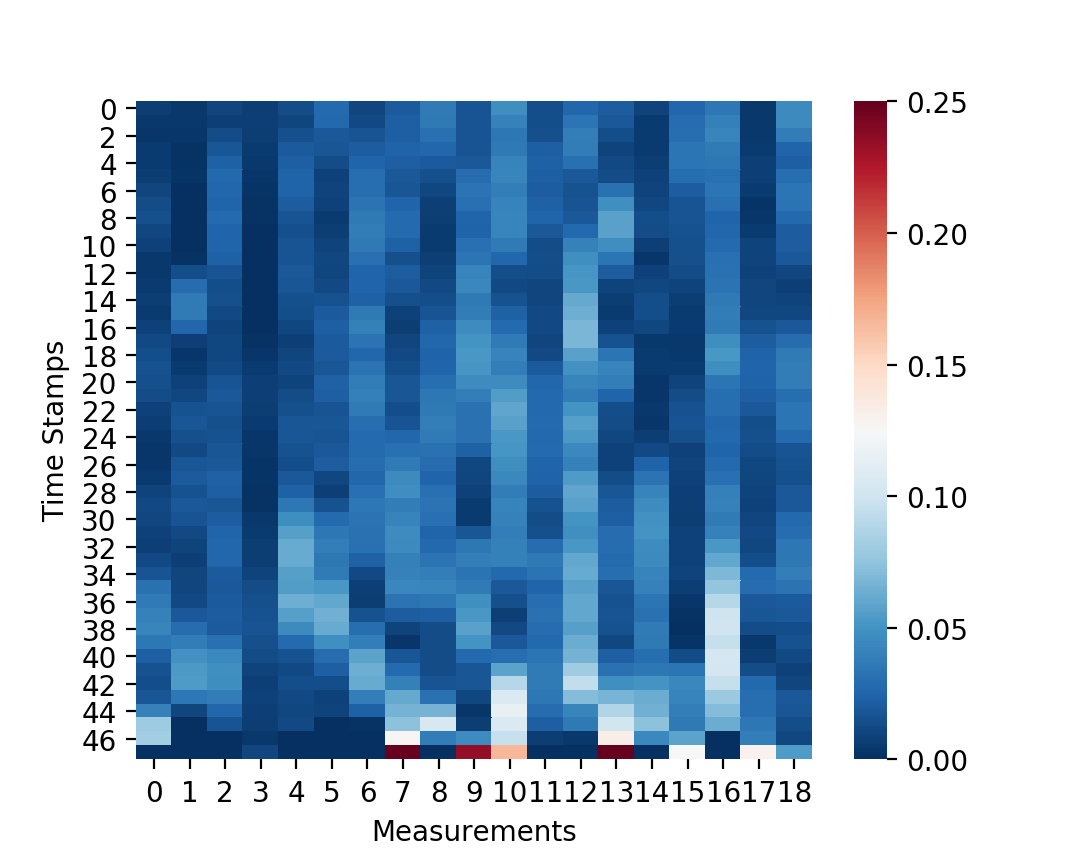} & 
        \includegraphics[width=0.25\textwidth]{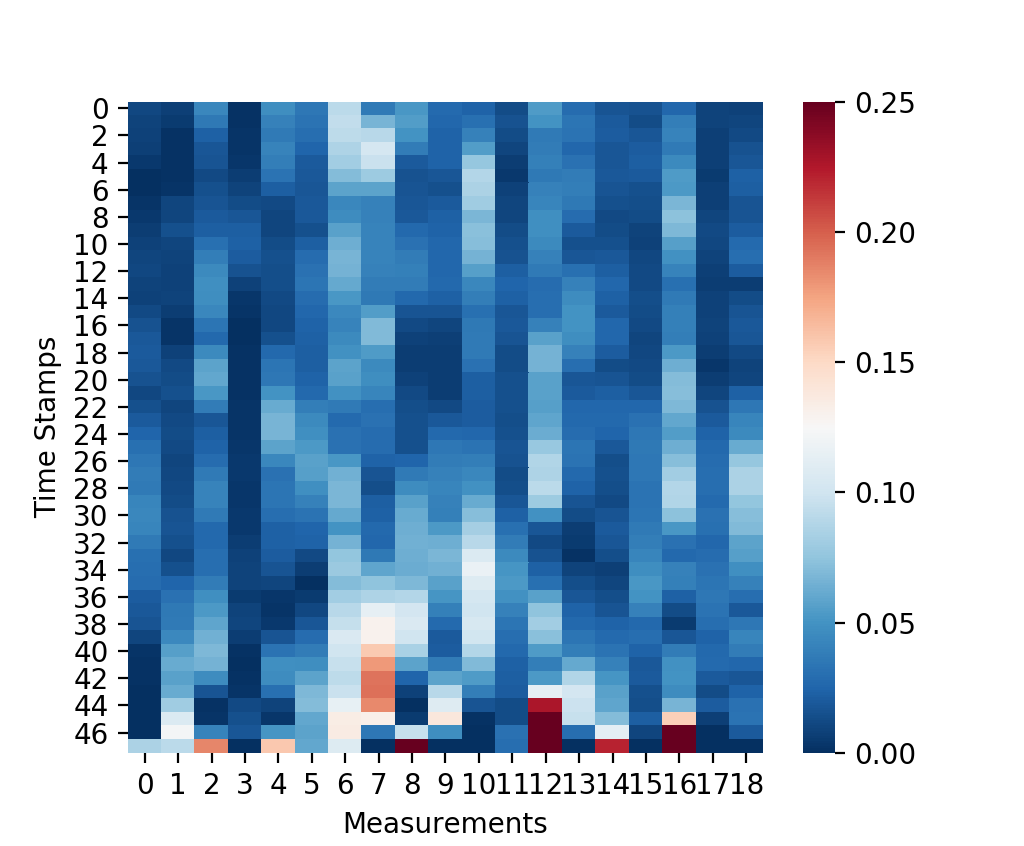} \\
        \multicolumn{2}{c}{(a) Maximum Perturbation}
        \\
        \includegraphics[width=0.25\textwidth]{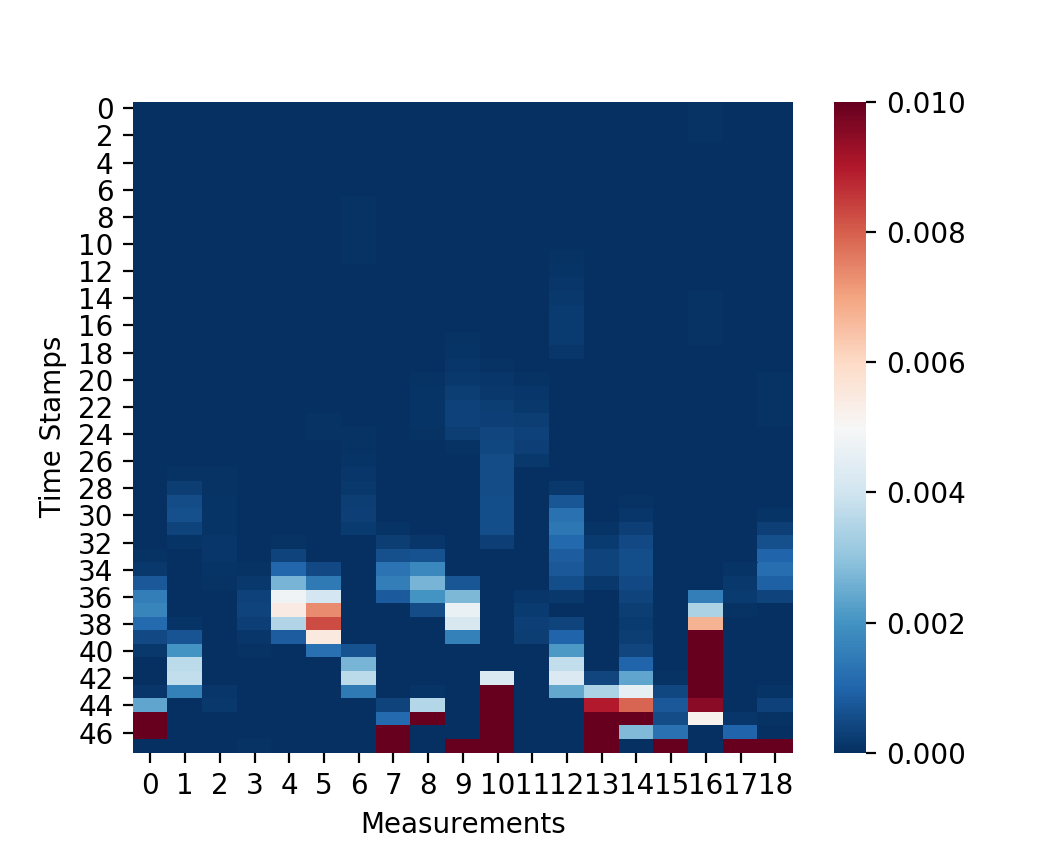} &
        \includegraphics[width=0.25\textwidth]{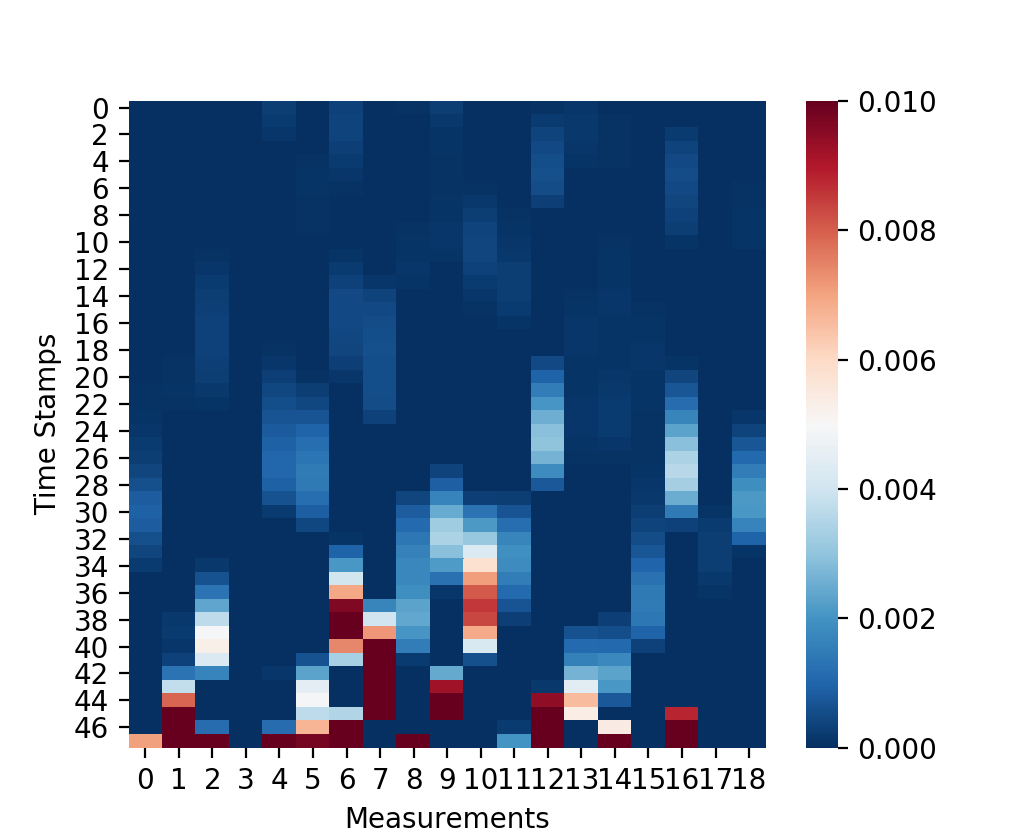} \\
        \multicolumn{2}{c}{(b) Average Perturbation}\\
        \includegraphics[width=0.25\textwidth]{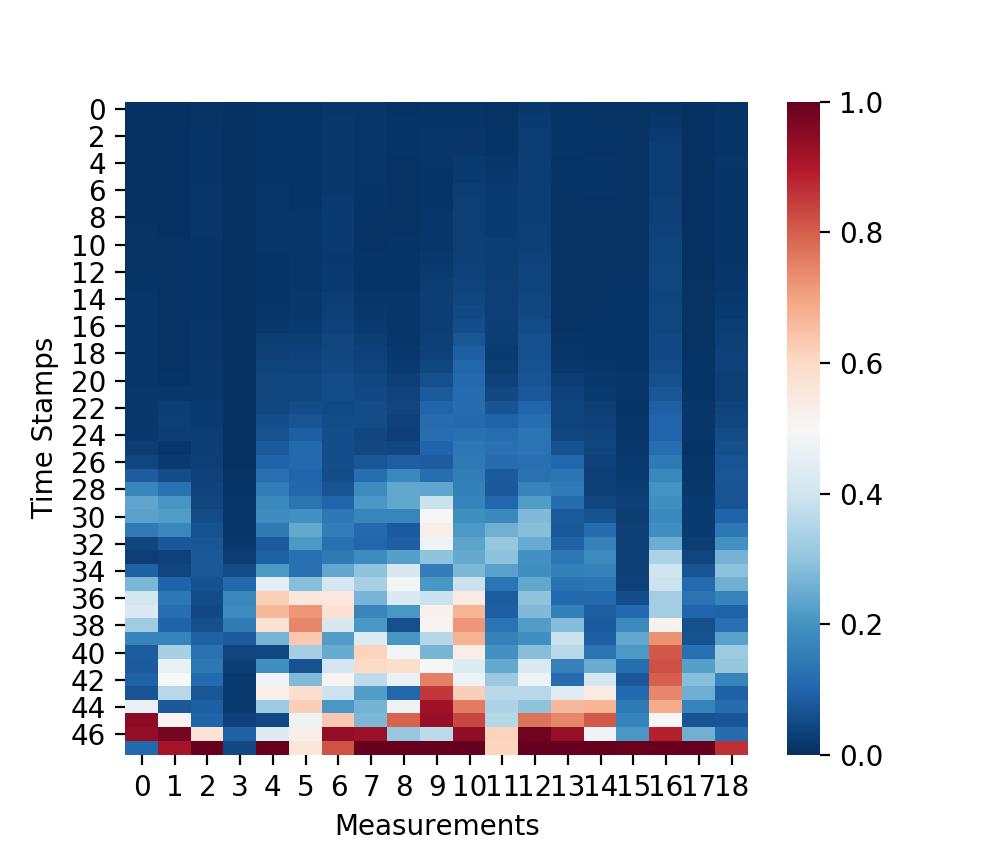} &
        \includegraphics[width=0.25\textwidth]{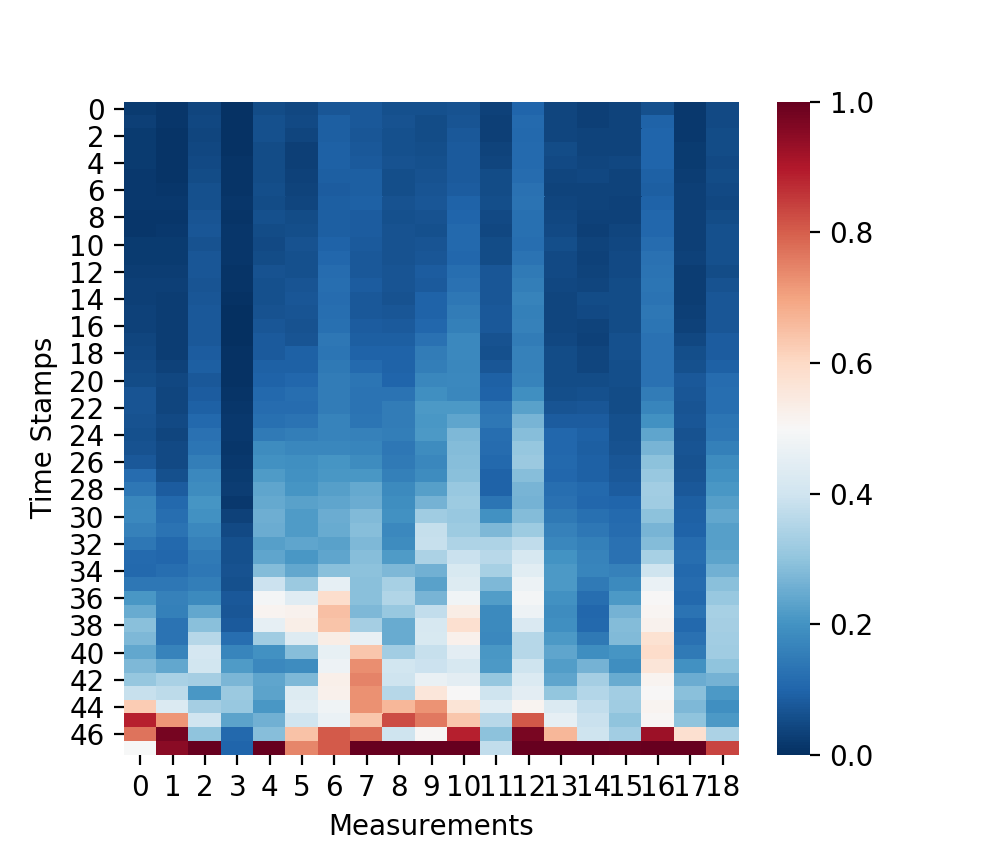} \\
        \multicolumn{2}{c}{(c) Perturbation Probability}
    \end{tabular}
    \vspace{-0.15in}
    \caption{Adversarial attacks (Left: 0 $\Rightarrow$ 1, Right: 1 $\Rightarrow$ 0) at population level .}\label{fig:pop-attack}
    \vspace{-0.15in}
\end{figure}

\noindent\textbf{Clinical Interpretation } By observing the differences in susceptibility to perturbation magnitude and direction across time, clinicians can determine which lab features are robust for consideration in personalized mortality risk assessment. For example, given the perturbation distribution (from an optimal adversarial record) for a patient, we can identify specific regions which warrant increased attention. Given the sample patient in Figure \ref{fig:p-01a}, we see that arterial carbon dioxide levels (PaCO2), Creatinine (Cre) and Sodium (Na) are more susceptible under the current prediction model. Specifically, we see that \emph{increase} in PaCO2, \emph{decrease} in Cre and Na levels render the original classifier susceptible to misclassification. In the reverse case, a comparison between \ref{fig:p-01a} and \ref{fig:p-10a} reveals that the direction of perturbation on PaCO2, Cre and Na are reversed to achieve misclassification from deceased to alive, which is consistent with our intuition. These results suggest that small errors in Cre, PaCO2 and Na measurements can introduce a significant vulnerability in the mortality assessment of these two sample patients.

Additionally, the direction of perturbation sensitivity can lead to varying interpretation at different time steps. %Although PaCO2 is particularly important at the final time step, PH, Albumin (Alb), bicarbonate (HCO3), Cre and Na levels exhibit susceptibility to perturbation at various time steps long before the prediction time. For example, Cre levels are susceptible to perturbation since the 28th time step while features such as PaCO2 and Glucose (Glc) are only relevant at the time of prediction. 
%From the clinical perspective, these results may guide the interpretation of feature importance at different time-steps. 
For example, we see from \ref{fig:p-01a} and \ref{fig:p-10a} that fluctuations in Na levels are susceptible to different directions of perturbation at various time points. At around 8 hours before prediction time (hours 38-44), overestimation in Na levels from laboratory tests renders the classifier susceptible to misclassification toward the alive label, while overestimation at prediction time (48th hour) may cause misclassification toward the deceased label. Perturbation matrix in this case may help guide interpretation of mortality risk with fluctuations in measurement at various time points for certain features.

%Additionally, we also see that time steps before the 24th hour, no features are susceptible to perturbation in both the alive and deceased misclassification cases. We may interpret this result as 
%The alive patient (0) can be misclassified to deceased (1) with maximum perturbation less than 10\%. 
%Typically, increasing carbon dioxide and decreasing creatinine and sodium are likely to cause a person to die. 
%We also find a reverse pattern by looking at patient \ref{fig:p-10a}, whose desired perturbations are opposite to patient \ref{fig:p-01a}, which is consistent with the the (0) to (1) case.

\begin{figure}[t!]
    \vspace{-0.1in}
    \centering
    \begin{subfigure}[b]{0.25\textwidth}
        \includegraphics[width=\textwidth]{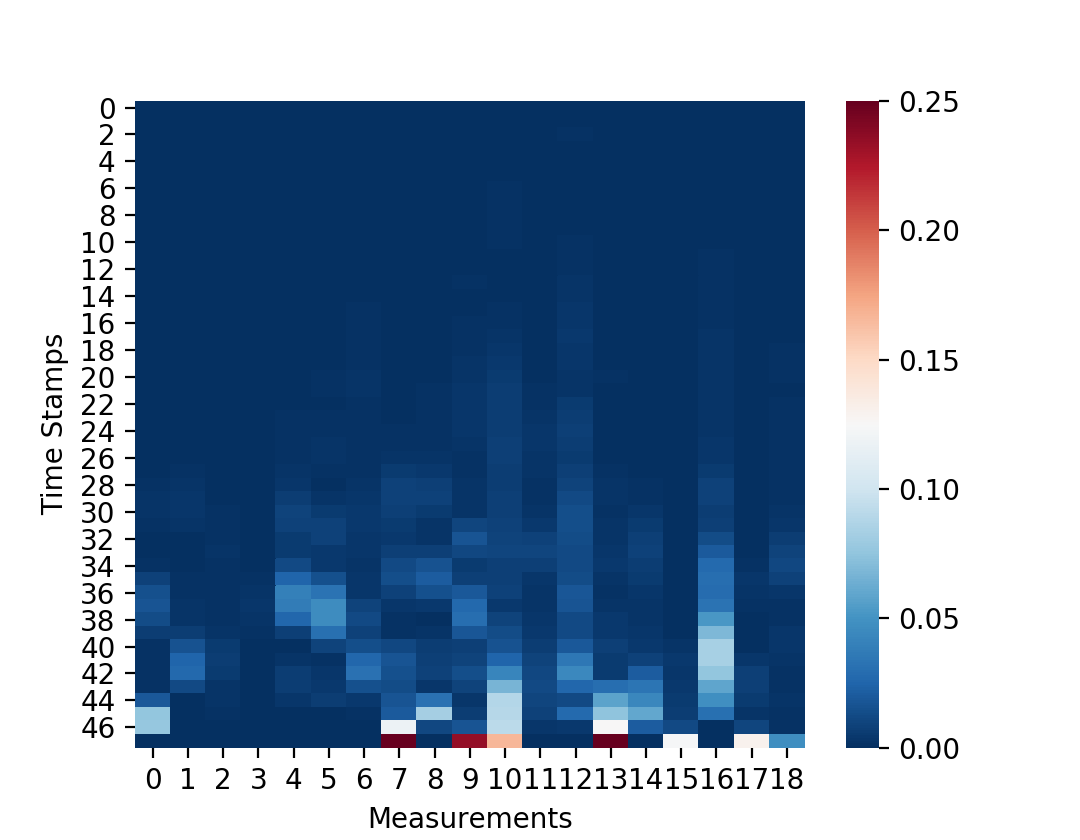}
        \caption{Time-Measurement sensitivity\\ score distribution.}
        \label{fig:ss-zero-1}
    \end{subfigure}
    ~ %add desired spacing between images, e. g. ~, \quad, \qquad, \hfill etc. 
    \begin{subfigure}[b]{0.2\textwidth}
        \includegraphics[width=\textwidth]{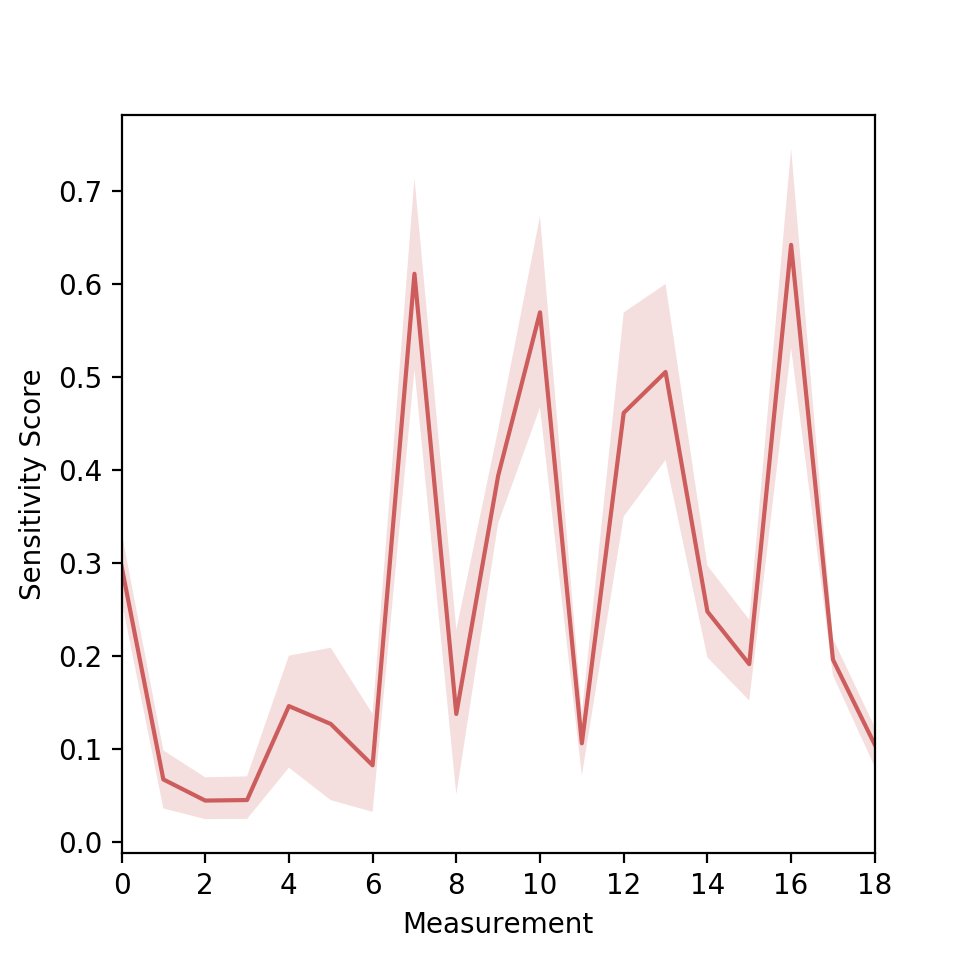}
        \caption{Measurement sensitivity score.}
        \label{fig:ss-zero-2}
    \end{subfigure}

    \vspace{-0.15in}
    \caption{Susceptibility at population level (0 to 1).}\label{fig:ss-zero}
    \vspace{-0.1in}
\end{figure}

\begin{figure}[t!]
    \vspace{-0.1in}
    \centering
    \begin{subfigure}[b]{0.21\textwidth}
        \includegraphics[width=\textwidth]{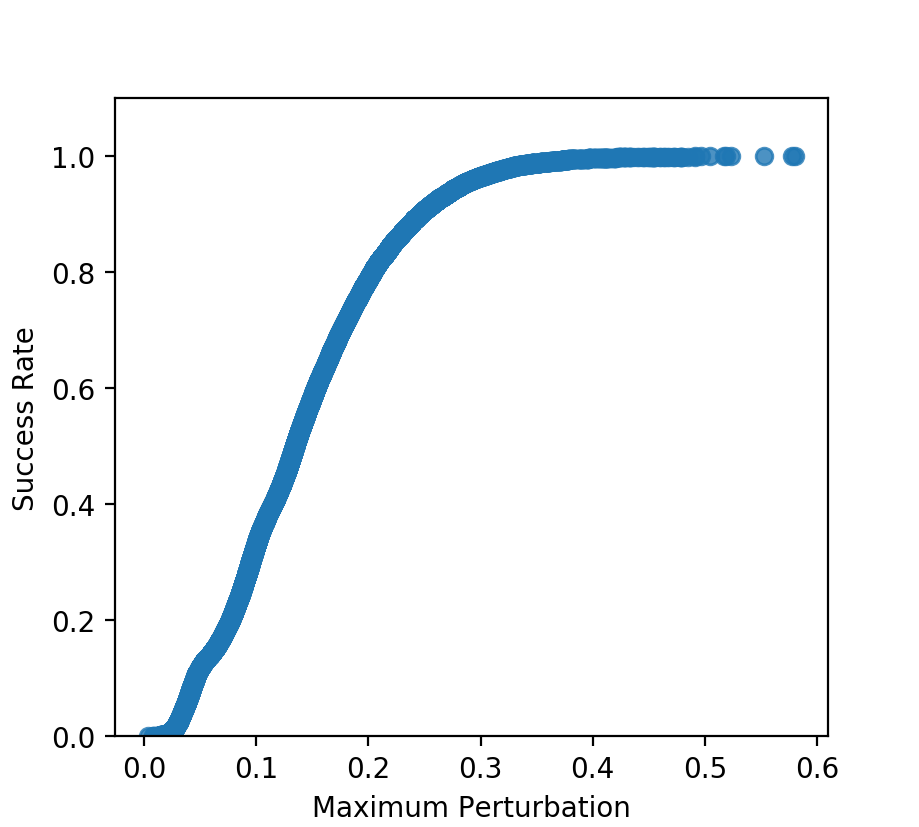}
        \caption{Success rate at different\\ maximum perturbation.}
        \label{fig:sr-zero-1}
    \end{subfigure}
    ~ %add desired spacing between images, e. g. ~, \quad, \qquad, \hfill etc. 
    \begin{subfigure}[b]{0.22\textwidth}
        \includegraphics[width=\textwidth]{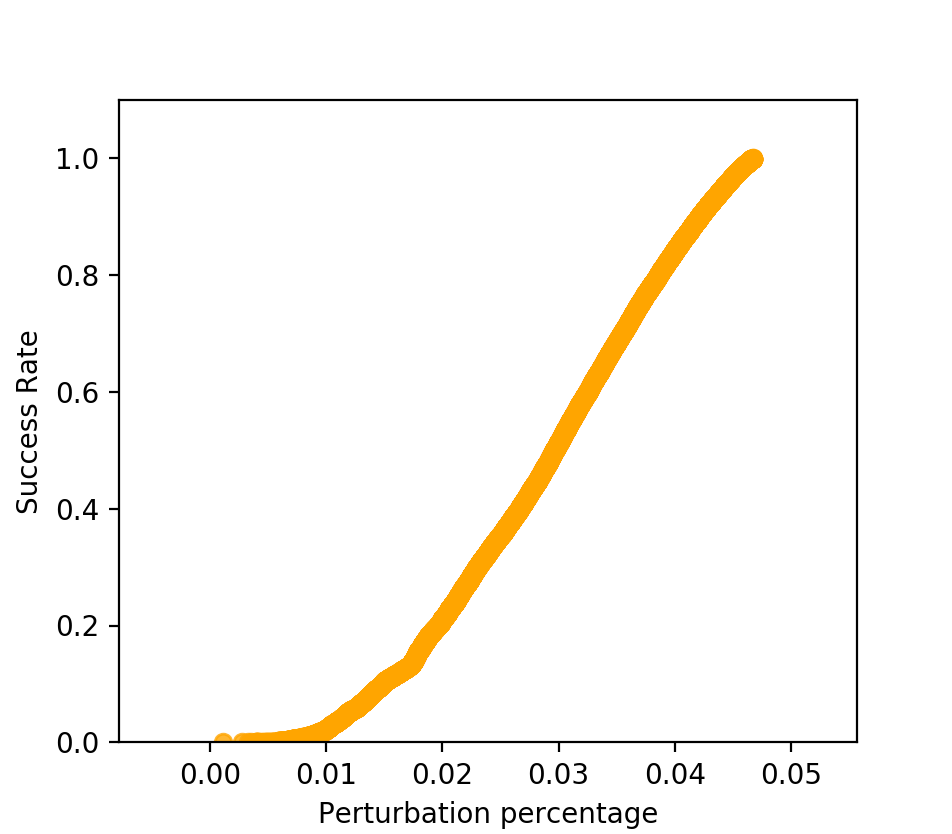}
        \caption{Average perturbation\\ percentage corresponds to \ref{fig:sr-zero-1}.}
        \label{fig:sr-zero-2}
    \end{subfigure}
    \vspace{-0.1in}
    \caption{Adversarial perturbation assessment.}\label{fig:sr-zero}
    \vspace{-0.2in}
\end{figure}

\vspace{+0.5em}
\noindent\textbf{Sensitivity at population Level.}
While patient-targeted attack reveals vulnerability signals at a personalized level, cohort or population-level analysis is needed to determine the vulnerability of diagnostic features over the entire EHR dataset. Therefore, we also adopt susceptibility score at the cohort level using predefined metrics. We first evaluate performance for a given fold and then extend to all the folds to check for consistency. 

\begin{figure*}[t!]
    \centering
    \begin{subfigure}[b]{0.33\textwidth}
        \includegraphics[width=\textwidth]{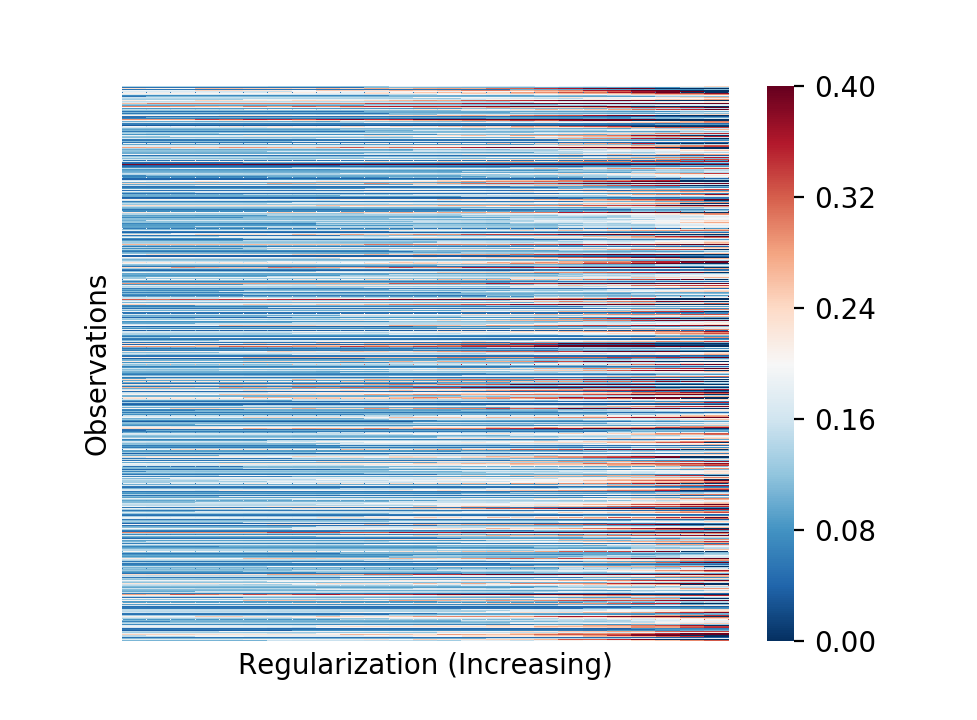}
        \caption{Maximum perturbation.}
        \label{fig:fold-1a}
    \end{subfigure}
    ~ %add desired spacing between images, e. g. ~, \quad, \qquad, \hfill etc. 
    \begin{subfigure}[b]{0.30\textwidth}
        \includegraphics[width=\textwidth]{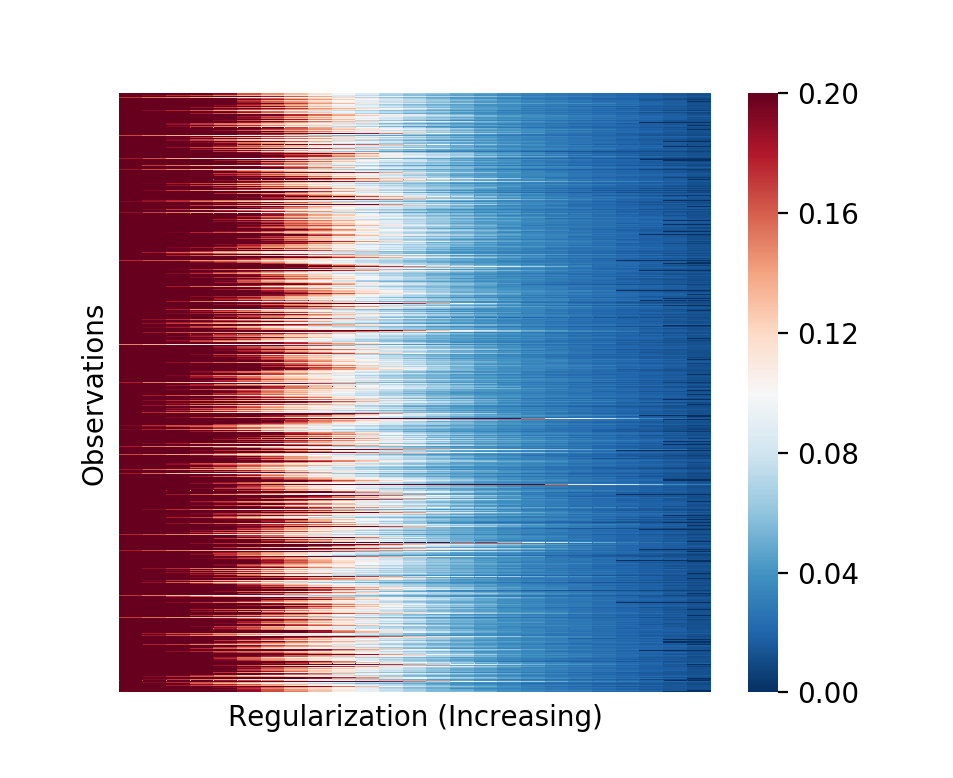}
        \caption{Perturbation percentage.}
        \label{fig:fold-1b}
    \end{subfigure}
    \begin{subfigure}[b]{0.30\textwidth}
    \includegraphics[width=\textwidth]{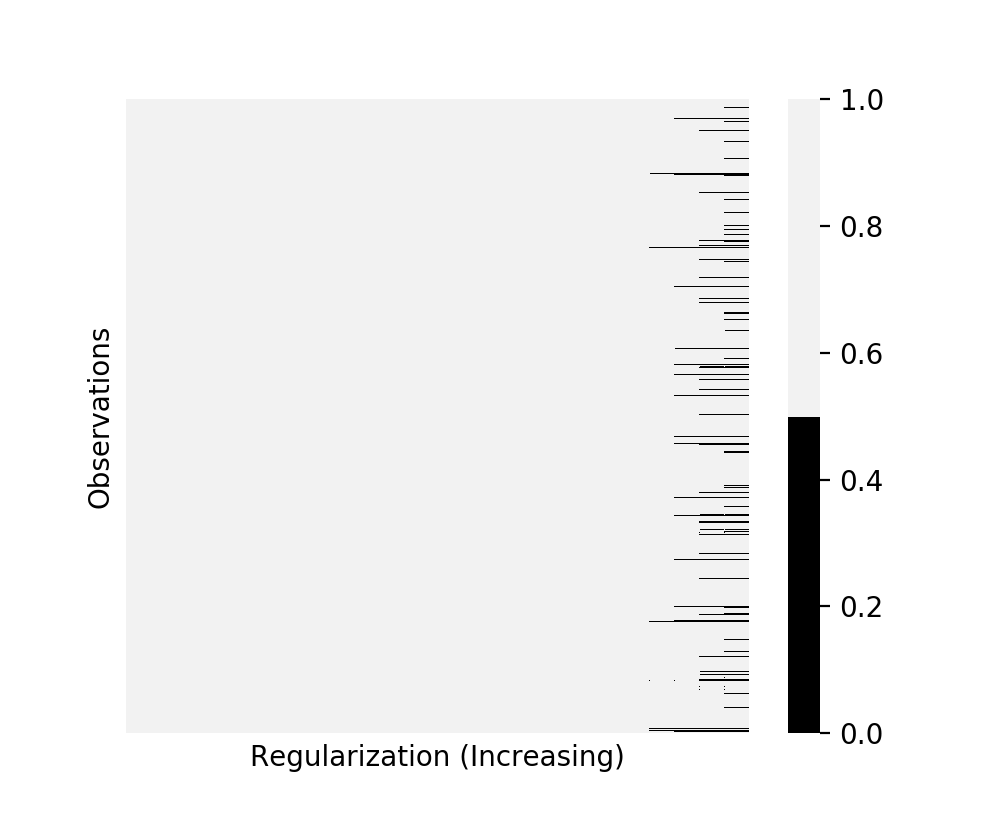}
    \caption{Success Indicator.}
    \label{fig:fold-1c}
\end{subfigure}
    \vspace{-0.1in}
    \caption{Adversarial attack for all patients in a random fold.}\label{fig:fold-1}
    \vspace{-0.1in}
\end{figure*}

\textit{Population-level trade-off.} Figure \ref{fig:fold-1} shows the maximum perturbation and percentage of perturbation under different sparsity control at population-level. We can see that as regularization increases, the maximum perturbation increases across all observations. Similarly, there is a more clear pattern for perturbation percentage with respect to regularization. A huge penalty would eventually encourage no spots to be changed and end up with failed adversarial records. Figure \ref{fig:fold-1c} indicates whether adversarial is successful, 0 for failure 1 for success respectively, across different regularizations. We can see that the attack generate success adversarial records most of the time.

\begin{figure}[!t]
    \vspace{-0.1in}
    \centering
    \begin{subfigure}[b]{0.26\textwidth}
        \includegraphics[width=\textwidth]{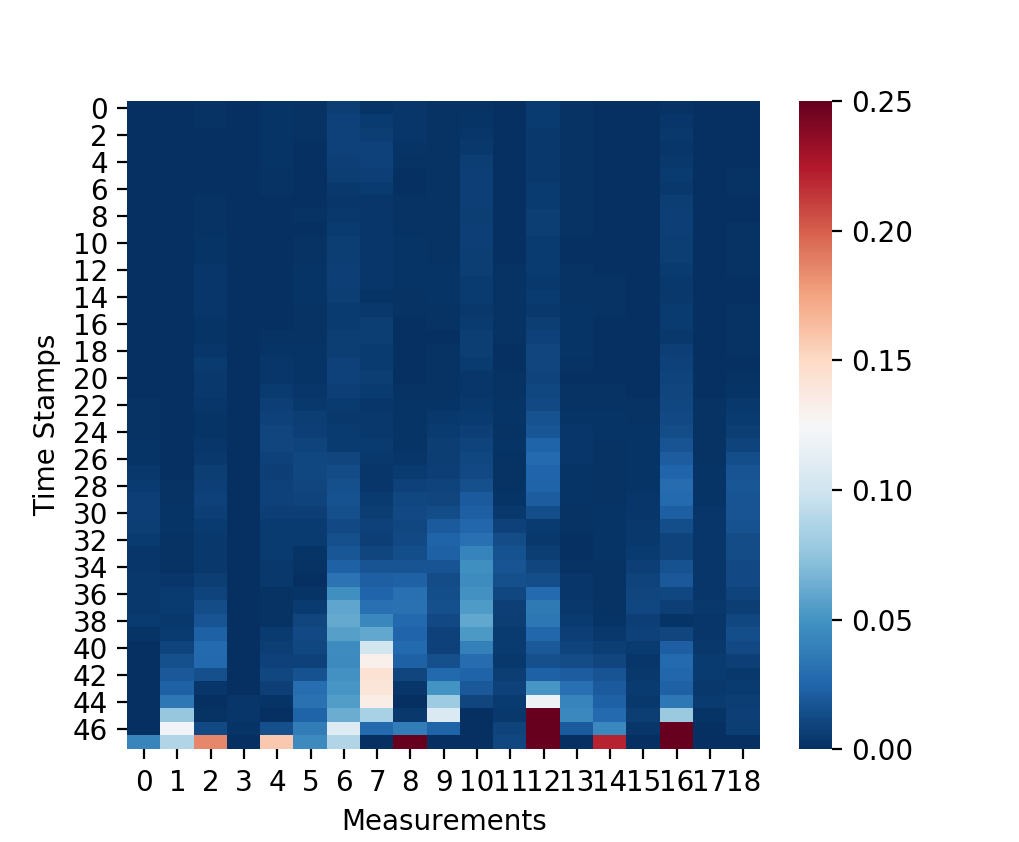}
        \caption{Time-Measurement score.}
        \label{fig:ss-one-1}
    \end{subfigure}
    ~ %add desired spacing between images, e. g. ~, \quad, \qquad, \hfill etc. 
    \begin{subfigure}[b]{0.23\textwidth}
        \includegraphics[width=\textwidth]{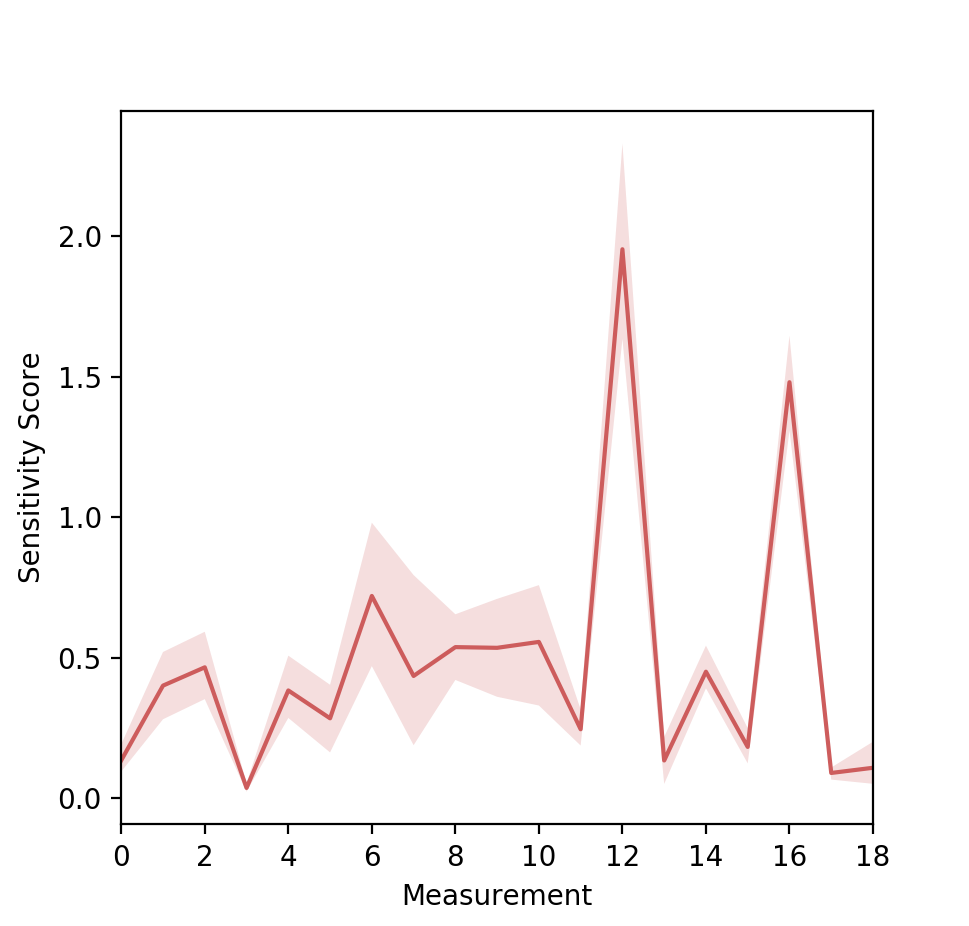}
        \caption{Measurement score.}
        \label{fig:ss-one-2}
    \end{subfigure}
    \vspace{-0.2in}
    \caption{Susceptibility at population level (1 to 0).}\label{fig:ss-one}
    \vspace{-0.1in}
\end{figure}

Figure \ref{fig:pop-attack} shows the global maximum perturbation (GMP), global average perturbation (GAP) and global perturbation probability (GPP) at each time-measurement grid for 0-1 attack ((left) and 1-0 attack (right) respectively. Figure \ref{fig:ss-zero-1} shows the susceptibility score distribution while considering all patients in a fold for 0-1 attack. We can see that measurement PaCO2, Albumin, Glc (7, 9 and 13) at prediction time are the most susceptible locations for mortality prediction model. Similar to the patient-targeted case, not all the measurements have perturbation susceptibility concentrated at the most recent time steps. We observe that measurements RR, SPO2, Na (4, 5 and 16) tends to be more susceptible at earlier time stamps. We repeat evaluation for all 5 folds and found that the results are quite similar. Considering all the time stamps, we obtain a susceptibility score for each measurement by summing the scores across the time axis. Figure \ref{fig:ss-zero-2} shows the average sensitivity score and 95\% confidence band for each measurement across all 5 folds. Same plots are made for 1-0 attack and can be found at Figure \ref{fig:ss-one}. Table \ref{rank-all} shows the ranking of sensitivity for all the measurements.
 
 \begin{table}[t!]
    \centering
    \caption{Rank of measurements based on susceptible score.}
    \vspace{-0.15in}
    \label{rank-all}\small
    \begin{tabular}{|c|c|c|c|c|}
        \hline
        \multirow{2}{*}{Rank} & \multicolumn{2}{c|}{0-1 Attack} & \multicolumn{2}{c|}{1-0 Attack} \\ \cline{2-5} 
        & Measurement & Susceptible Score & Measurement & Susceptible Score \\ \hline\hline
        1                     & Na          & 0.64 (0.13)        & Cre         & 1.95 (0.46)        \\ \hline
        2                     & PaCO2       & 0.61 (0.13)        & Na          & 1.48 (0.21)        \\ \hline
        3                     & HCO3        & 0.57 (0.13)        & Lactate     & 0.72 (0.33)        \\ \hline
        4                     & Glc         & 0.51 (0.12)        & HCO3        & 0.56 (0.27)        \\ \hline
        5                     & Cre         & 0.46 (0.15)        & PH          & 0.54 (0.15)        \\ \hline
        6                     & Albumin     & 0.39 (0.07)        & Albumin     & 0.54 (0.23)        \\ \hline
        7                     & HR          & 0.30 0.05)        & DBP         & 0.47 (0.15)        \\ \hline
        8                     & Mg          & 0.25 (0.06)        & Mg          & 0.45 (0.11)        \\ \hline
        9                     & BUN         & 0.20 (0.03)        & PaCO2       & 0.43 (0.39)        \\ \hline
        10                    & K           & 0.19 (0.06)        & SBP         & 0.40 (0.15)        \\ \hline
        11                    & RR          & 0.15 (0.08)        & RR          & 0.38 (0.14)        \\ \hline
        12                    & PH          & 0.14 (0.12)        & SPO2        & 0.28 (0.16)        \\ \hline
        13                    & SPO2        & 0.13 (0.11)        & Ca          & 0.25 (0.08)        \\ \hline
        14                    & Ca          & 0.11 (0.04)        & K           & 0.18 (0.08)        \\ \hline
        15                    & Platelets   & 0.10 (0.03)        & HR          & 0.13 (0.06)        \\ \hline
        16                    & Lactate     & 0.08 (0.07)        & Glc         & 0.13 (0.11)        \\ \hline
        17                    & SBP         & 0.07 (0.04)        & Platelets   & 0.11 (0.10)        \\ \hline
        18                    & TEMP        & 0.05 (0.03)        & BUN         & 0.09 (0.03)        \\ \hline
        19                    & DBP         & 0.04 (0.03)        & TEMP        & 0.04 (0.02)        \\ \hline
    \end{tabular}
    \vspace{-0.2in}
\end{table}

\noindent\textbf{Clinical Interpretation }
Similar to the patient-level patterns, we also observe differences in perturbation sparsity and magnitude across various features at the population-level. Figure \ref{fig:pop-attack} (c) shows the probability that a perturbation at certain time steps would lead to misclassification from alive to deceased. This figure can be interpreted as the distribution of susceptibility of each feature across time over the entire EHR dataset. Cold-spots in this grid relate to low probabilities of attack, signaling to the clinician that features associated with those areas are more robust to attack. On the other hand, Figure \ref{fig:pop-attack} (b) illustrates the average perturbation over the EHR population for each feature across time. In this grid, we are more concerned about the hot-spots, as cold-spots alone do not rule-out the possibility of LSTM-attack since they only indicate small \emph{magnitude} of perturbations. However, hot-spots in this grid, indicate preferential spots of LSTM-attack, which rules-in highly susceptible features of the EHR population for misclassification. Figure \ref{fig:pop-attack} (a) illustrates the element-wise max perturbation for each feature across all time-steps in the database. As seen from \ref{eqt:obj}, LSTM-attacks are regularized by the magnitude of the perturbation. In the case where the original LSTM classifier is uncertain of the classification decision, the associated data points are closer to the decision boundary, requiring smaller perturbation to achieve misclassification. In the opposite case where data points are further away from the classifier decision boundary, more perturbation is required to cause misclassification. Thus, the max-perturbation grid shown in \ref{fig:pop-attack} (a) represents the average LSTM prediction certainty over the EHR record samples; a max-perturbation grid that is dominated by cold-spots reflects a classifier that is mostly uncertain of its predictions, yielding to small perturbations required for misclassification. A max-perturbation grid that is hot-spot dominant indicates a classifier that is highly certain of its predictions, as max-thresholds of perturbation required is higher for most of its samples. %From the sparsity level of perturbation probability, we see that the recency preference of LSTM-attack is evident for almost all features, but certain lab features, i.e., Cre, PH and Albumin levels are attacked across time for numerous patients. This finding suggests that small perturbations in these features matter across various time-steps despite changes in 
%At the population-level, we observe clear differences between the max perturbation and average perturbation magnitudes as shown in \ref{fig:pop-1a} and \ref{fig:pop-1b}. In figure \ref{fig:pop-1b}

Additionally, these results may be used to provide clinical decision support in determining the optimal rate of sampling for individual clinical features. For example, while vital signs (features 0-5) are obtained at all times for each patient admitted to the urgent care, laboratory tests such as comprehensive metabolic panels (CMP) are much more costly to obtain, especially for patients who require a long length of stays ~\cite{hsia2014variation}.
%Urgent care settings are typically equipped to provide instant feedback for these clinical features due to their ubiquity and low cost. However, .
Using our attack framework, a clinician can opt to perform an initial prediction at the beginning of an admission to gauge mortality risk of a patient, and then determine the sampling rate for certain laboratory tests depending on the susceptibility score of the clinical feature to perturbations across time. 
For example, Table \ref{rank-all} can give clinicians an initial interpretation of the validity of the mortality risk obtained from a prediction model at admission time, while Figure \ref{fig:ss-zero} may inform when and how frequently subsequent laboratory tests should be done to update mortality risk evaluation. By sampling at an optimal rate which adjusts for perturbation of measurements across time, clinicians can potentially save cost associated with frequent use of expensive diagnostic tests without sacrificing the accuracy of clinical assessment.

\subsection{Adversarial assessment.}
While our ultimate goal is to utilize attack results to detect susceptibility locations of clinical records,  we do not enforce strict constraints on the attacks. We assess our adversarial attack by success rate achieved at different maximum perturbation across all patients and the corresponding perturbation percentage at each stage. Results are shown in Figure \ref{fig:sr-zero}. We can see that more than half of the patients can be perturbed with maximum perturbation less than 0.15 by only changing 3\% of the record locations for 0-1 attack. The 1-0 direction is relatively harder to attack compared to the other.

\begin{figure}[t!]
    \vspace{-0.1in}
    \centering
    \begin{subfigure}[b]{0.24\textwidth}
        \includegraphics[width=\textwidth]{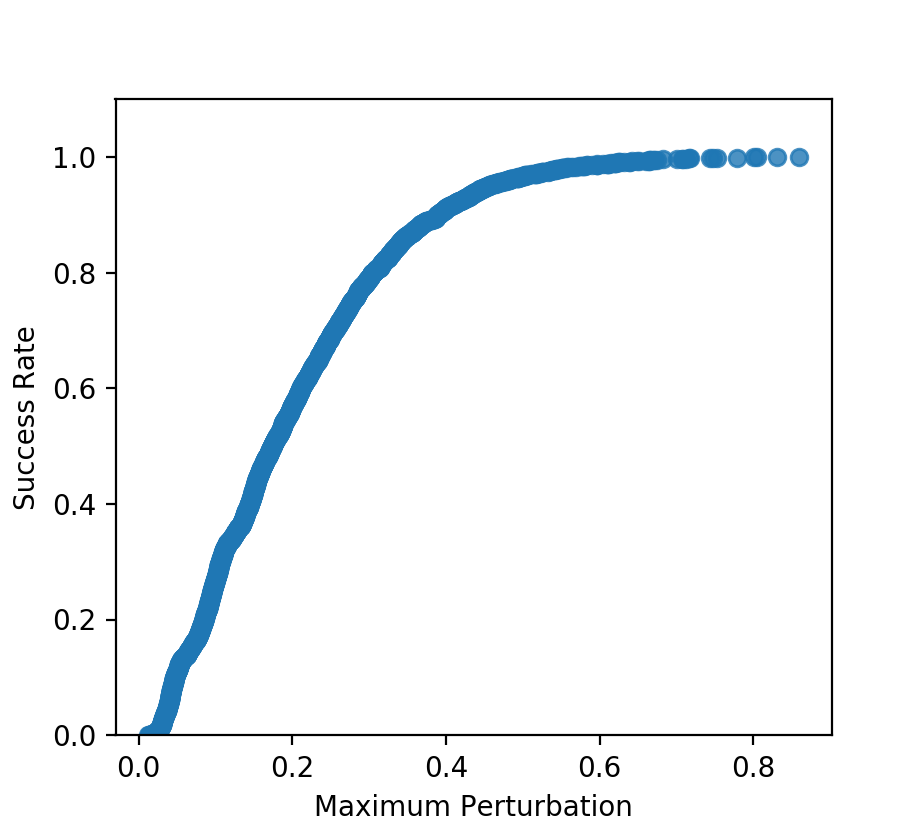}
        \caption{Success rate at different\\ maximum perturbation.}
        \label{fig:sr-one-1}
    \end{subfigure}
    ~ %add desired spacing between images, e. g. ~, \quad, \qquad, \hfill etc. 
    \begin{subfigure}[b]{0.23\textwidth}
        \includegraphics[width=\textwidth]{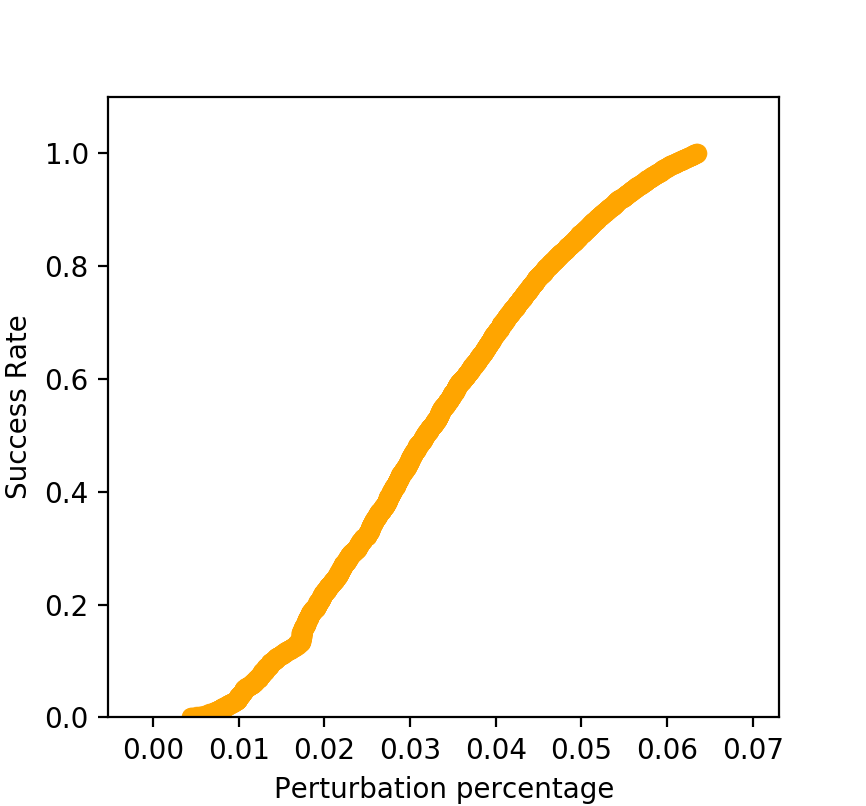}
        \caption{Average perturbation\\ percentage corresponds to \ref{fig:sr-one-1}.}
        \label{fig:sr-one-2}
    \end{subfigure}
    \vspace{-0.1in}
    \caption{Adversarial perturbation assessment (1 to 0).}\label{fig:sr-one}
    \vspace{-0.2in}
\end{figure}

%1. individual level: heat map for 1 ob, combination for 1 ob, clinical interpretation 2. one-fold level: MP, AP, PN of all sparsity control;  MP, AP, PN for all observations; qualitative number of perturbation results. 3. cross-fold level: whether consistent 4. extend to other diseases, conclusion

\section{Conclusion}\label{sec:conclusion}
In this paper, we proposed an efficient and effective framework that identifies susceptible locations in medical records utilizing adversarial attacks on deep predictive models. Our results demonstrated the vulnerability of deep models, where more than half of the patients can be successfully attacked by changing  only 3\% of the record locations with maximum perturbation less than 0.15 and average perturbation less than 0.02. The proposed screening approach can detect susceptible events and locations at both patient and population level, providing valuable information and assistance for clinical professionals. In addition, the framework can be easily extend to other predictive modeling on time series data.

%
%\section*{Appendix}
%\input{appendix}

\begin{acks}
This research is supported in part by 
National Science Foundation under Grant
IIS-1565596 (JZ), IIS-1615597 (JZ), IIS-1650723 (FW) and IIS-1716432 (FW). and the Office of Naval Research under
grant number N00014-14-1-0631 (JZ) and N00014-17-1-2265 (JZ).
\end{acks}

\bibliographystyle{ACM-Reference-Format}
\bibliography{reference}


\begin{thebibliography}{40}

%%% the Web address.

\ifx \showCODEN    \undefined \def \showCODEN     #1{\unskip}     \fi
\ifx \showDOI      \undefined \def \showDOI       #1{#1}\fi
\ifx \showISBNx    \undefined \def \showISBNx     #1{\unskip}     \fi
\ifx \showISBNxiii \undefined \def \showISBNxiii  #1{\unskip}     \fi
\ifx \showISSN     \undefined \def \showISSN      #1{\unskip}     \fi
\ifx \showLCCN     \undefined \def \showLCCN      #1{\unskip}     \fi
\ifx \shownote     \undefined \def \shownote      #1{#1}          \fi
\ifx \showarticletitle \undefined \def \showarticletitle #1{#1}   \fi
\ifx \showURL      \undefined \def \showURL       {\relax}        \fi
% invisible to TeX
\providecommand\bibfield[2]{#2}
\providecommand\bibinfo[2]{#2}
\providecommand\natexlab[1]{#1}
\providecommand\showeprint[2][]{arXiv:#2}

\bibitem[\protect\citeauthoryear{Alvin~Rajkomar}{Alvin~Rajkomar}{2018}]%
        {Rajkomar2018scalable}
\bibfield{author}{\bibinfo{person}{Eyal Oren et.~al Alvin~Rajkomar}.}
  \bibinfo{year}{2018}\natexlab{}.
\newblock \showarticletitle{Scalable and accurate deep learning for electronic
  health records}.
\newblock \bibinfo{journal}{\emph{arXiv preprint arXiv:1801.07860}}
  (\bibinfo{year}{2018}).
\newblock


\bibitem[\protect\citeauthoryear{Anderson, Kharkar, Filar, and Roth}{Anderson
  et~al\mbox{.}}{2017}]%
        {anderson2017evading}
\bibfield{author}{\bibinfo{person}{Hyrum~S Anderson}, \bibinfo{person}{Anant
  Kharkar}, \bibinfo{person}{Bobby Filar}, {and} \bibinfo{person}{Phil Roth}.}
  \bibinfo{year}{2017}\natexlab{}.
\newblock \showarticletitle{Evading machine learning malware detection}.
\newblock \bibinfo{journal}{\emph{Black Hat}} (\bibinfo{year}{2017}).
\newblock


\bibitem[\protect\citeauthoryear{Anderson, Woodbridge, and Filar}{Anderson
  et~al\mbox{.}}{2016}]%
        {anderson2016deepdga}
\bibfield{author}{\bibinfo{person}{Hyrum~S Anderson}, \bibinfo{person}{Jonathan
  Woodbridge}, {and} \bibinfo{person}{Bobby Filar}.}
  \bibinfo{year}{2016}\natexlab{}.
\newblock \showarticletitle{DeepDGA: Adversarially-Tuned Domain Generation and
  Detection}. In \bibinfo{booktitle}{\emph{Proceedings of the 2016 ACM Workshop
  on Artificial Intelligence and Security}}. ACM, \bibinfo{pages}{13--21}.
\newblock


\bibitem[\protect\citeauthoryear{Baytas, Xiao, Zhang, Wang, Jain, and
  Zhou}{Baytas et~al\mbox{.}}{2017}]%
        {baytas2017patient}
\bibfield{author}{\bibinfo{person}{Inci~M Baytas}, \bibinfo{person}{Cao Xiao},
  \bibinfo{person}{Xi Zhang}, \bibinfo{person}{Fei Wang},
  \bibinfo{person}{Anil~K Jain}, {and} \bibinfo{person}{Jiayu Zhou}.}
  \bibinfo{year}{2017}\natexlab{}.
\newblock \showarticletitle{Patient subtyping via time-aware LSTM networks}. In
  \bibinfo{booktitle}{\emph{Proceedings of the 23rd ACM SIGKDD International
  Conference on Knowledge Discovery and Data Mining}}. ACM,
  \bibinfo{pages}{65--74}.
\newblock


\bibitem[\protect\citeauthoryear{Beck and Teboulle}{Beck and Teboulle}{2009}]%
        {beck2009fast}
\bibfield{author}{\bibinfo{person}{Amir Beck} {and} \bibinfo{person}{Marc
  Teboulle}.} \bibinfo{year}{2009}\natexlab{}.
\newblock \showarticletitle{A fast iterative shrinkage-thresholding algorithm
  for linear inverse problems}.
\newblock \bibinfo{journal}{\emph{SIAM journal on imaging sciences}}
  \bibinfo{volume}{2}, \bibinfo{number}{1} (\bibinfo{year}{2009}),
  \bibinfo{pages}{183--202}.
\newblock


\bibitem[\protect\citeauthoryear{Carlini, Katz, Barrett, and Dill}{Carlini
  et~al\mbox{.}}{2017}]%
        {carlini2017ground}
\bibfield{author}{\bibinfo{person}{Nicholas Carlini}, \bibinfo{person}{Guy
  Katz}, \bibinfo{person}{Clark Barrett}, {and} \bibinfo{person}{David~L
  Dill}.} \bibinfo{year}{2017}\natexlab{}.
\newblock \showarticletitle{Ground-Truth Adversarial Examples}.
\newblock \bibinfo{journal}{\emph{arXiv preprint arXiv:1709.10207}}
  (\bibinfo{year}{2017}).
\newblock


\bibitem[\protect\citeauthoryear{Carlini and Wagner}{Carlini and
  Wagner}{2017}]%
        {carlini2017towards}
\bibfield{author}{\bibinfo{person}{Nicholas Carlini} {and}
  \bibinfo{person}{David Wagner}.} \bibinfo{year}{2017}\natexlab{}.
\newblock \showarticletitle{Towards evaluating the robustness of neural
  networks}. In \bibinfo{booktitle}{\emph{Security and Privacy (SP), 2017 IEEE
  Symposium on}}. IEEE, \bibinfo{pages}{39--57}.
\newblock


\bibitem[\protect\citeauthoryear{Che, Kale, Li, Bahadori, and Liu}{Che
  et~al\mbox{.}}{2015}]%
        {che2015deep}
\bibfield{author}{\bibinfo{person}{Zhengping Che}, \bibinfo{person}{David
  Kale}, \bibinfo{person}{Wenzhe Li}, \bibinfo{person}{Mohammad~Taha Bahadori},
  {and} \bibinfo{person}{Yan Liu}.} \bibinfo{year}{2015}\natexlab{}.
\newblock \showarticletitle{Deep computational phenotyping}. In
  \bibinfo{booktitle}{\emph{Proceedings of the 21th ACM SIGKDD International
  Conference on Knowledge Discovery and Data Mining}}. ACM,
  \bibinfo{pages}{507--516}.
\newblock


\bibitem[\protect\citeauthoryear{Che, Purushotham, Khemani, and Liu}{Che
  et~al\mbox{.}}{2016}]%
        {che2016interpretable}
\bibfield{author}{\bibinfo{person}{Zhengping Che}, \bibinfo{person}{Sanjay
  Purushotham}, \bibinfo{person}{Robinder Khemani}, {and} \bibinfo{person}{Yan
  Liu}.} \bibinfo{year}{2016}\natexlab{}.
\newblock \showarticletitle{Interpretable deep models for icu outcome
  prediction}. In \bibinfo{booktitle}{\emph{AMIA Annual Symposium
  Proceedings}}, Vol.~\bibinfo{volume}{2016}. American Medical Informatics
  Association, \bibinfo{pages}{371}.
\newblock


\bibitem[\protect\citeauthoryear{Chen, Sharma, Zhang, Yi, and Hsieh}{Chen
  et~al\mbox{.}}{2017a}]%
        {chen2017ead}
\bibfield{author}{\bibinfo{person}{Pin-Yu Chen}, \bibinfo{person}{Yash Sharma},
  \bibinfo{person}{Huan Zhang}, \bibinfo{person}{Jinfeng Yi}, {and}
  \bibinfo{person}{Cho-Jui Hsieh}.} \bibinfo{year}{2017}\natexlab{a}.
\newblock \showarticletitle{EAD: Elastic-Net Attacks to Deep Neural Networks
  via Adversarial Examples}.
\newblock \bibinfo{journal}{\emph{arXiv preprint arXiv:1709.04114}}
  (\bibinfo{year}{2017}).
\newblock


\bibitem[\protect\citeauthoryear{Chen, Zhang, Sharma, Yi, and Hsieh}{Chen
  et~al\mbox{.}}{2017b}]%
        {chen2017zoo}
\bibfield{author}{\bibinfo{person}{Pin-Yu Chen}, \bibinfo{person}{Huan Zhang},
  \bibinfo{person}{Yash Sharma}, \bibinfo{person}{Jinfeng Yi}, {and}
  \bibinfo{person}{Cho-Jui Hsieh}.} \bibinfo{year}{2017}\natexlab{b}.
\newblock \showarticletitle{Zoo: Zeroth order optimization based black-box
  attacks to deep neural networks without training substitute models}. In
  \bibinfo{booktitle}{\emph{Proceedings of the 10th ACM Workshop on Artificial
  Intelligence and Security}}. ACM, \bibinfo{pages}{15--26}.
\newblock


\bibitem[\protect\citeauthoryear{Cho, Van~Merri{\"e}nboer, Gulcehre, Bahdanau,
  Bougares, Schwenk, and Bengio}{Cho et~al\mbox{.}}{2014}]%
        {cho2014learning}
\bibfield{author}{\bibinfo{person}{Kyunghyun Cho}, \bibinfo{person}{Bart
  Van~Merri{\"e}nboer}, \bibinfo{person}{Caglar Gulcehre},
  \bibinfo{person}{Dzmitry Bahdanau}, \bibinfo{person}{Fethi Bougares},
  \bibinfo{person}{Holger Schwenk}, {and} \bibinfo{person}{Yoshua Bengio}.}
  \bibinfo{year}{2014}\natexlab{}.
\newblock \showarticletitle{Learning phrase representations using RNN
  encoder-decoder for statistical machine translation}.
\newblock \bibinfo{journal}{\emph{arXiv preprint arXiv:1406.1078}}
  (\bibinfo{year}{2014}).
\newblock


\bibitem[\protect\citeauthoryear{Choi, Bahadori, Schuetz, Stewart, and
  Sun}{Choi et~al\mbox{.}}{2016a}]%
        {choi2016doctor}
\bibfield{author}{\bibinfo{person}{Edward Choi}, \bibinfo{person}{Mohammad~Taha
  Bahadori}, \bibinfo{person}{Andy Schuetz}, \bibinfo{person}{Walter~F
  Stewart}, {and} \bibinfo{person}{Jimeng Sun}.}
  \bibinfo{year}{2016}\natexlab{a}.
\newblock \showarticletitle{Doctor ai: Predicting clinical events via recurrent
  neural networks}. In \bibinfo{booktitle}{\emph{Machine Learning for
  Healthcare Conference}}. \bibinfo{pages}{301--318}.
\newblock


\bibitem[\protect\citeauthoryear{Choi, Bahadori, Searles, Coffey, Thompson,
  Bost, Tejedor-Sojo, and Sun}{Choi et~al\mbox{.}}{2016b}]%
        {choi2016multi}
\bibfield{author}{\bibinfo{person}{Edward Choi}, \bibinfo{person}{Mohammad~Taha
  Bahadori}, \bibinfo{person}{Elizabeth Searles}, \bibinfo{person}{Catherine
  Coffey}, \bibinfo{person}{Michael Thompson}, \bibinfo{person}{James Bost},
  \bibinfo{person}{Javier Tejedor-Sojo}, {and} \bibinfo{person}{Jimeng Sun}.}
  \bibinfo{year}{2016}\natexlab{b}.
\newblock \showarticletitle{Multi-layer representation learning for medical
  concepts}. In \bibinfo{booktitle}{\emph{Proceedings of the 22nd ACM SIGKDD
  International Conference on Knowledge Discovery and Data Mining}}. ACM,
  \bibinfo{pages}{1495--1504}.
\newblock


\bibitem[\protect\citeauthoryear{Goodfellow, Shlens, and Szegedy}{Goodfellow
  et~al\mbox{.}}{2014}]%
        {goodfellow2014explaining}
\bibfield{author}{\bibinfo{person}{Ian~J Goodfellow}, \bibinfo{person}{Jonathon
  Shlens}, {and} \bibinfo{person}{Christian Szegedy}.}
  \bibinfo{year}{2014}\natexlab{}.
\newblock \showarticletitle{Explaining and harnessing adversarial examples}.
\newblock \bibinfo{journal}{\emph{arXiv preprint arXiv:1412.6572}}
  (\bibinfo{year}{2014}).
\newblock


\bibitem[\protect\citeauthoryear{Grosse, Papernot, Manoharan, Backes, and
  McDaniel}{Grosse et~al\mbox{.}}{2017}]%
        {grosse2017adversarial}
\bibfield{author}{\bibinfo{person}{Kathrin Grosse}, \bibinfo{person}{Nicolas
  Papernot}, \bibinfo{person}{Praveen Manoharan}, \bibinfo{person}{Michael
  Backes}, {and} \bibinfo{person}{Patrick McDaniel}.}
  \bibinfo{year}{2017}\natexlab{}.
\newblock \showarticletitle{Adversarial examples for malware detection}. In
  \bibinfo{booktitle}{\emph{European Symposium on Research in Computer
  Security}}. Springer, \bibinfo{pages}{62--79}.
\newblock


\bibitem[\protect\citeauthoryear{Harutyunyan, Khachatrian, Kale, and
  Galstyan}{Harutyunyan et~al\mbox{.}}{2017}]%
        {harutyunyan2017multitask}
\bibfield{author}{\bibinfo{person}{Hrayr Harutyunyan}, \bibinfo{person}{Hrant
  Khachatrian}, \bibinfo{person}{David~C Kale}, {and} \bibinfo{person}{Aram
  Galstyan}.} \bibinfo{year}{2017}\natexlab{}.
\newblock \showarticletitle{Multitask Learning and Benchmarking with Clinical
  Time Series Data}.
\newblock \bibinfo{journal}{\emph{arXiv preprint arXiv:1703.07771}}
  (\bibinfo{year}{2017}).
\newblock


\bibitem[\protect\citeauthoryear{Hochreiter and Schmidhuber}{Hochreiter and
  Schmidhuber}{1997}]%
        {hochreiter1997long}
\bibfield{author}{\bibinfo{person}{Sepp Hochreiter} {and}
  \bibinfo{person}{J{\"u}rgen Schmidhuber}.} \bibinfo{year}{1997}\natexlab{}.
\newblock \showarticletitle{Long short-term memory}.
\newblock \bibinfo{journal}{\emph{Neural computation}} \bibinfo{volume}{9},
  \bibinfo{number}{8} (\bibinfo{year}{1997}), \bibinfo{pages}{1735--1780}.
\newblock


\bibitem[\protect\citeauthoryear{Hsia, Antwi, and Nath}{Hsia
  et~al\mbox{.}}{2014}]%
        {hsia2014variation}
\bibfield{author}{\bibinfo{person}{Renee~Y Hsia}, \bibinfo{person}{Yaa~Akosa
  Antwi}, {and} \bibinfo{person}{Julia~P Nath}.}
  \bibinfo{year}{2014}\natexlab{}.
\newblock \showarticletitle{Variation in charges for 10 common blood tests in
  California hospitals: a cross-sectional analysis}.
\newblock \bibinfo{journal}{\emph{BMJ open}} \bibinfo{volume}{4},
  \bibinfo{number}{8} (\bibinfo{year}{2014}), \bibinfo{pages}{e005482}.
\newblock


\bibitem[\protect\citeauthoryear{Hu and Tan}{Hu and Tan}{2017}]%
        {hu2017generating}
\bibfield{author}{\bibinfo{person}{Weiwei Hu} {and} \bibinfo{person}{Ying
  Tan}.} \bibinfo{year}{2017}\natexlab{}.
\newblock \showarticletitle{Generating Adversarial Malware Examples for
  Black-Box Attacks Based on GAN}.
\newblock \bibinfo{journal}{\emph{arXiv preprint arXiv:1702.05983}}
  (\bibinfo{year}{2017}).
\newblock


\bibitem[\protect\citeauthoryear{Jia and Liang}{Jia and Liang}{2017}]%
        {jia2017adversarial}
\bibfield{author}{\bibinfo{person}{Robin Jia} {and} \bibinfo{person}{Percy
  Liang}.} \bibinfo{year}{2017}\natexlab{}.
\newblock \showarticletitle{Adversarial examples for evaluating reading
  comprehension systems}.
\newblock \bibinfo{journal}{\emph{arXiv preprint arXiv:1707.07328}}
  (\bibinfo{year}{2017}).
\newblock


\bibitem[\protect\citeauthoryear{Johnson, Pollard, Shen, Li-wei, Feng,
  Ghassemi, Moody, Szolovits, Celi, and Mark}{Johnson et~al\mbox{.}}{2016}]%
        {johnson2016mimic}
\bibfield{author}{\bibinfo{person}{Alistair~EW Johnson}, \bibinfo{person}{Tom~J
  Pollard}, \bibinfo{person}{Lu Shen}, \bibinfo{person}{H~Lehman Li-wei},
  \bibinfo{person}{Mengling Feng}, \bibinfo{person}{Mohammad Ghassemi},
  \bibinfo{person}{Benjamin Moody}, \bibinfo{person}{Peter Szolovits},
  \bibinfo{person}{Leo~Anthony Celi}, {and} \bibinfo{person}{Roger~G Mark}.}
  \bibinfo{year}{2016}\natexlab{}.
\newblock \showarticletitle{MIMIC-III, a freely accessible critical care
  database}.
\newblock \bibinfo{journal}{\emph{Scientific data}}  \bibinfo{volume}{3}
  (\bibinfo{year}{2016}), \bibinfo{pages}{160035}.
\newblock


\bibitem[\protect\citeauthoryear{Li, Monroe, and Jurafsky}{Li
  et~al\mbox{.}}{2016}]%
        {li2016understanding}
\bibfield{author}{\bibinfo{person}{Jiwei Li}, \bibinfo{person}{Will Monroe},
  {and} \bibinfo{person}{Dan Jurafsky}.} \bibinfo{year}{2016}\natexlab{}.
\newblock \showarticletitle{Understanding Neural Networks through
  Representation Erasure}.
\newblock \bibinfo{journal}{\emph{arXiv preprint arXiv:1612.08220}}
  (\bibinfo{year}{2016}).
\newblock


\bibitem[\protect\citeauthoryear{Lipton, Kale, Elkan, and Wetzel}{Lipton
  et~al\mbox{.}}{2015}]%
        {lipton2015learning}
\bibfield{author}{\bibinfo{person}{Zachary~C Lipton}, \bibinfo{person}{David~C
  Kale}, \bibinfo{person}{Charles Elkan}, {and} \bibinfo{person}{Randall
  Wetzel}.} \bibinfo{year}{2015}\natexlab{}.
\newblock \showarticletitle{Learning to diagnose with LSTM recurrent neural
  networks}.
\newblock \bibinfo{journal}{\emph{arXiv preprint arXiv:1511.03677}}
  (\bibinfo{year}{2015}).
\newblock


\bibitem[\protect\citeauthoryear{Liu, Chen, Liu, and Song}{Liu
  et~al\mbox{.}}{2016}]%
        {liu2016delving}
\bibfield{author}{\bibinfo{person}{Yanpei Liu}, \bibinfo{person}{Xinyun Chen},
  \bibinfo{person}{Chang Liu}, {and} \bibinfo{person}{Dawn Song}.}
  \bibinfo{year}{2016}\natexlab{}.
\newblock \showarticletitle{Delving into transferable adversarial examples and
  black-box attacks}.
\newblock \bibinfo{journal}{\emph{arXiv preprint arXiv:1611.02770}}
  (\bibinfo{year}{2016}).
\newblock


\bibitem[\protect\citeauthoryear{Miotto, Li, Kidd, and Dudley}{Miotto
  et~al\mbox{.}}{2016}]%
        {miotto2016deep}
\bibfield{author}{\bibinfo{person}{Riccardo Miotto}, \bibinfo{person}{Li Li},
  \bibinfo{person}{Brian~A Kidd}, {and} \bibinfo{person}{Joel~T Dudley}.}
  \bibinfo{year}{2016}\natexlab{}.
\newblock \showarticletitle{Deep patient: an unsupervised representation to
  predict the future of patients from the electronic health records}.
\newblock \bibinfo{journal}{\emph{Scientific reports}}  \bibinfo{volume}{6}
  (\bibinfo{year}{2016}), \bibinfo{pages}{26094}.
\newblock


\bibitem[\protect\citeauthoryear{Moosavi-Dezfooli, Fawzi, Fawzi, and
  Frossard}{Moosavi-Dezfooli et~al\mbox{.}}{2016b}]%
        {moosavi2016universal}
\bibfield{author}{\bibinfo{person}{Seyed-Mohsen Moosavi-Dezfooli},
  \bibinfo{person}{Alhussein Fawzi}, \bibinfo{person}{Omar Fawzi}, {and}
  \bibinfo{person}{Pascal Frossard}.} \bibinfo{year}{2016}\natexlab{b}.
\newblock \showarticletitle{Universal adversarial perturbations}.
\newblock \bibinfo{journal}{\emph{arXiv preprint arXiv:1610.08401}}
  (\bibinfo{year}{2016}).
\newblock


\bibitem[\protect\citeauthoryear{Moosavi-Dezfooli, Fawzi, and
  Frossard}{Moosavi-Dezfooli et~al\mbox{.}}{2016a}]%
        {moosavi2016deepfool}
\bibfield{author}{\bibinfo{person}{Seyed-Mohsen Moosavi-Dezfooli},
  \bibinfo{person}{Alhussein Fawzi}, {and} \bibinfo{person}{Pascal Frossard}.}
  \bibinfo{year}{2016}\natexlab{a}.
\newblock \showarticletitle{Deepfool: a simple and accurate method to fool deep
  neural networks}. In \bibinfo{booktitle}{\emph{Proceedings of the IEEE
  Conference on Computer Vision and Pattern Recognition}}.
  \bibinfo{pages}{2574--2582}.
\newblock


\bibitem[\protect\citeauthoryear{Nguyen, Yosinski, and Clune}{Nguyen
  et~al\mbox{.}}{2015}]%
        {nguyen2015deep}
\bibfield{author}{\bibinfo{person}{Anh Nguyen}, \bibinfo{person}{Jason
  Yosinski}, {and} \bibinfo{person}{Jeff Clune}.}
  \bibinfo{year}{2015}\natexlab{}.
\newblock \showarticletitle{Deep neural networks are easily fooled: High
  confidence predictions for unrecognizable images}. In
  \bibinfo{booktitle}{\emph{Proceedings of the IEEE Conference on Computer
  Vision and Pattern Recognition}}. \bibinfo{pages}{427--436}.
\newblock


\bibitem[\protect\citeauthoryear{Nguyen, Tran, Wickramasinghe, and
  Venkatesh}{Nguyen et~al\mbox{.}}{2017}]%
        {nguyen2017mathtt}
\bibfield{author}{\bibinfo{person}{Phuoc Nguyen}, \bibinfo{person}{Truyen
  Tran}, \bibinfo{person}{Nilmini Wickramasinghe}, {and}
  \bibinfo{person}{Svetha Venkatesh}.} \bibinfo{year}{2017}\natexlab{}.
\newblock \showarticletitle{Deepr: A Convolutional Net for Medical Records}.
\newblock \bibinfo{journal}{\emph{IEEE journal of biomedical and health
  informatics}} \bibinfo{volume}{21}, \bibinfo{number}{1}
  (\bibinfo{year}{2017}), \bibinfo{pages}{22--30}.
\newblock


\bibitem[\protect\citeauthoryear{Papernot, McDaniel, Jha, Fredrikson, Celik,
  and Swami}{Papernot et~al\mbox{.}}{2016a}]%
        {papernot2016limitations}
\bibfield{author}{\bibinfo{person}{Nicolas Papernot}, \bibinfo{person}{Patrick
  McDaniel}, \bibinfo{person}{Somesh Jha}, \bibinfo{person}{Matt Fredrikson},
  \bibinfo{person}{Z~Berkay Celik}, {and} \bibinfo{person}{Ananthram Swami}.}
  \bibinfo{year}{2016}\natexlab{a}.
\newblock \showarticletitle{The limitations of deep learning in adversarial
  settings}. In \bibinfo{booktitle}{\emph{Security and Privacy (EuroS\&P), 2016
  IEEE European Symposium on}}. IEEE, \bibinfo{pages}{372--387}.
\newblock


\bibitem[\protect\citeauthoryear{Papernot, McDaniel, Swami, and
  Harang}{Papernot et~al\mbox{.}}{2016b}]%
        {papernot2016crafting}
\bibfield{author}{\bibinfo{person}{Nicolas Papernot}, \bibinfo{person}{Patrick
  McDaniel}, \bibinfo{person}{Ananthram Swami}, {and} \bibinfo{person}{Richard
  Harang}.} \bibinfo{year}{2016}\natexlab{b}.
\newblock \showarticletitle{Crafting adversarial input sequences for recurrent
  neural networks}. In \bibinfo{booktitle}{\emph{Military Communications
  Conference, MILCOM 2016-2016 IEEE}}. IEEE, \bibinfo{pages}{49--54}.
\newblock


\bibitem[\protect\citeauthoryear{Pham, Tran, Phung, and Venkatesh}{Pham
  et~al\mbox{.}}{2016}]%
        {pham2016deepcare}
\bibfield{author}{\bibinfo{person}{Trang Pham}, \bibinfo{person}{Truyen Tran},
  \bibinfo{person}{Dinh Phung}, {and} \bibinfo{person}{Svetha Venkatesh}.}
  \bibinfo{year}{2016}\natexlab{}.
\newblock \showarticletitle{Deepcare: A deep dynamic memory model for
  predictive medicine}. In \bibinfo{booktitle}{\emph{Pacific-Asia Conference on
  Knowledge Discovery and Data Mining}}. Springer, \bibinfo{pages}{30--41}.
\newblock


\bibitem[\protect\citeauthoryear{Rozsa, Rudd, and Boult}{Rozsa
  et~al\mbox{.}}{2016}]%
        {rozsa2016adversarial}
\bibfield{author}{\bibinfo{person}{Andras Rozsa}, \bibinfo{person}{Ethan~M
  Rudd}, {and} \bibinfo{person}{Terrance~E Boult}.}
  \bibinfo{year}{2016}\natexlab{}.
\newblock \showarticletitle{Adversarial diversity and hard positive
  generation}. In \bibinfo{booktitle}{\emph{Proceedings of the IEEE Conference
  on Computer Vision and Pattern Recognition Workshops}}.
  \bibinfo{pages}{25--32}.
\newblock


\bibitem[\protect\citeauthoryear{Sabour, Cao, Faghri, and Fleet}{Sabour
  et~al\mbox{.}}{2015}]%
        {sabour2015adversarial}
\bibfield{author}{\bibinfo{person}{Sara Sabour}, \bibinfo{person}{Yanshuai
  Cao}, \bibinfo{person}{Fartash Faghri}, {and} \bibinfo{person}{David~J
  Fleet}.} \bibinfo{year}{2015}\natexlab{}.
\newblock \showarticletitle{Adversarial manipulation of deep representations}.
\newblock \bibinfo{journal}{\emph{arXiv preprint arXiv:1511.05122}}
  (\bibinfo{year}{2015}).
\newblock


\bibitem[\protect\citeauthoryear{Shickel, Tighe, Bihorac, and Rashidi}{Shickel
  et~al\mbox{.}}{2017}]%
        {shickel2017deep}
\bibfield{author}{\bibinfo{person}{Benjamin Shickel},
  \bibinfo{person}{Patrick~James Tighe}, \bibinfo{person}{Azra Bihorac}, {and}
  \bibinfo{person}{Parisa Rashidi}.} \bibinfo{year}{2017}\natexlab{}.
\newblock \showarticletitle{Deep EHR: A Survey of Recent Advances in Deep
  Learning Techniques for Electronic Health Record (EHR) Analysis}.
\newblock \bibinfo{journal}{\emph{IEEE Journal of Biomedical and Health
  Informatics}} (\bibinfo{year}{2017}).
\newblock


\bibitem[\protect\citeauthoryear{Szegedy, Zaremba, Sutskever, Bruna, Erhan,
  Goodfellow, and Fergus}{Szegedy et~al\mbox{.}}{2013}]%
        {szegedy2013intriguing}
\bibfield{author}{\bibinfo{person}{Christian Szegedy},
  \bibinfo{person}{Wojciech Zaremba}, \bibinfo{person}{Ilya Sutskever},
  \bibinfo{person}{Joan Bruna}, \bibinfo{person}{Dumitru Erhan},
  \bibinfo{person}{Ian Goodfellow}, {and} \bibinfo{person}{Rob Fergus}.}
  \bibinfo{year}{2013}\natexlab{}.
\newblock \showarticletitle{Intriguing properties of neural networks}.
\newblock \bibinfo{journal}{\emph{arXiv preprint arXiv:1312.6199}}
  (\bibinfo{year}{2013}).
\newblock


\bibitem[\protect\citeauthoryear{Wang, Lee, Hu, Sun, and Ebadollahi}{Wang
  et~al\mbox{.}}{2012}]%
        {wang2012towards}
\bibfield{author}{\bibinfo{person}{Fei Wang}, \bibinfo{person}{Noah Lee},
  \bibinfo{person}{Jianying Hu}, \bibinfo{person}{Jimeng Sun}, {and}
  \bibinfo{person}{Shahram Ebadollahi}.} \bibinfo{year}{2012}\natexlab{}.
\newblock \showarticletitle{Towards heterogeneous temporal clinical event
  pattern discovery: a convolutional approach}. In
  \bibinfo{booktitle}{\emph{Proceedings of the 18th ACM SIGKDD international
  conference on Knowledge discovery and data mining}}. ACM,
  \bibinfo{pages}{453--461}.
\newblock


\bibitem[\protect\citeauthoryear{Zhao, Dua, and Singh}{Zhao
  et~al\mbox{.}}{2017}]%
        {zhao2017generating}
\bibfield{author}{\bibinfo{person}{Zhengli Zhao}, \bibinfo{person}{Dheeru Dua},
  {and} \bibinfo{person}{Sameer Singh}.} \bibinfo{year}{2017}\natexlab{}.
\newblock \showarticletitle{Generating Natural Adversarial Examples}.
\newblock \bibinfo{journal}{\emph{arXiv preprint arXiv:1710.11342}}
  (\bibinfo{year}{2017}).
\newblock


\bibitem[\protect\citeauthoryear{Zhou, Wang, Hu, and Ye}{Zhou
  et~al\mbox{.}}{2014}]%
        {zhou2014micro}
\bibfield{author}{\bibinfo{person}{Jiayu Zhou}, \bibinfo{person}{Fei Wang},
  \bibinfo{person}{Jianying Hu}, {and} \bibinfo{person}{Jieping Ye}.}
  \bibinfo{year}{2014}\natexlab{}.
\newblock \showarticletitle{From micro to macro: data driven phenotyping by
  densification of longitudinal electronic medical records}. In
  \bibinfo{booktitle}{\emph{Proceedings of the 20th ACM SIGKDD international
  conference on Knowledge discovery and data mining}}. ACM,
  \bibinfo{pages}{135--144}.
\newblock


\end{thebibliography}

\end{document}